\documentclass[11pt,letterpaper]{article}
\usepackage{caption}
\usepackage{subcaption}
\usepackage{fullpage}
\usepackage{times}
\usepackage{array}
\usepackage{ragged2e}
\usepackage{multirow}
\usepackage{fancyhdr, amssymb}
\usepackage[ruled,vlined]{algorithm2e}
\include{pythonlisting}
\usepackage[dvipsnames]{xcolor}
\usepackage[margin = 1 in]{geometry}
\usepackage{graphicx}
\usepackage{outlines}
\usepackage{amsmath}

\usepackage{graphicx}
\usepackage{dirtytalk}

\usepackage[english]{babel} 
\usepackage{blindtext}
\usepackage{xltabular}
\usepackage{arydshln} 
\usepackage{wrapfig}
\usepackage{enumitem}
\usepackage{floatrow}
\usepackage{xfrac}
\usepackage[font={footnotesize}, labelfont={bf}, labelsep=space, justification=justified, skip=0pt]{caption}
\usepackage[hang,flushmargin]{footmisc} 
\usepackage[colorlinks=true, allcolors=blue]{hyperref}

\usepackage[
backend=biber,
style=numeric-comp,
sorting=none,
maxnames = 10
]{biblatex}
\usepackage{cleveref} 

\newcommand{\cmt}[1]{} 

\newcommand{\ub}{\boldsymbol{u}}
\newcommand{\tb}{\boldsymbol{t}}
\newcommand{\xb}{\boldsymbol{x}}
\newcommand{\zb}{\boldsymbol{z}}
\newcommand{\hb}{\boldsymbol{h}}
\newcommand{\Thetab}{\boldsymbol{\Theta}}

\newcommand{\kmean}{$k$-mean }
\newcommand{\kmeans}{$k$-means }
\newcommand{\manifoldlearning}{manifold learning }
\newcommand{\Manifoldlearning}{Manifold learning }

\usepackage{titling}
\thanksheadextra{}{}
\thanksheadextra{}{} 
\usepackage{lipsum} 

\title{Unsupervised Anomaly Detection via Nonlinear Manifold Learning}
\date{\vspace{-5ex}}
\usepackage{authblk}
\author[1]{Amin Yousefpour}
\author[1]{Mehdi Shishehbor}
\author[1]{Zahra Zanjani Foumani}
\author[1]{Ramin Bostanabad \thanks{ Corresponding Author: Raminb@uci.edu \\\href{https://gitlab.com/S3anaz/multi-fidelity-cost-aware-bayesian-optimization/-/tree/main/}{Gitlab repository: https://}}}
\affil[1]{Department of Mechanical and Aerospace Engineering, University of California, Irvine}
\begin{document}

    \pagenumbering{arabic}
    \sloppy
    \maketitle
    
    \noindent \textbf{Abstract}\\
Anomalies are samples that significantly deviate from the rest of the data and their detection plays a major role in building machine learning models that can be reliably used in applications such as data-driven design and novelty detection. 
The majority of existing anomaly detection methods either are exclusively developed for (semi) supervised settings, or provide poor performance in unsupervised applications where there is no training data with labeled anomalous samples.
To bridge this research gap, we introduce a robust, efficient, and interpretable methodology based on nonlinear manifold learning to detect anomalies in unsupervised settings. 
The essence of our approach is to learn a low-dimensional and interpretable latent representation (aka manifold) for all the data points such that normal samples are automatically clustered together and hence can be easily and robustly identified. We learn this low-dimensional manifold by designing a learning algorithm that leverages either a latent map Gaussian process (LMGP) or a deep autoencoder (AE). Our LMGP-based approach, in particular, provides a probabilistic perspective on the learning task and is ideal for high-dimensional applications with scarce data. 
We demonstrate the superior performance of our approach over existing technologies via multiple analytic examples and real-world datasets. 

    \section{Introduction} \label{sec: intro}

Anomalies, i.e., observations that deviate significantly from the majority of instances, ubiquitously exist in real-world datasets. These samples can dramatically affect the performance of machine learning (ML) models \cite{edgeworth1887xli} and arise for a multitude of reasons such as errors in the data generation/recording mechanism or existence of uncharacterized features \cite{chandola2009anomaly}. For instance, factors such as faulty equipment, environmental conditions, or changes in the manufacturing process \cite{garmaroodi2020detection} can lead to anomalous samples in industrial applications. Failure to detect these anomalies results in inaccurate predictions and decreased reliability in data-driven design applications \cite{skomedal2020much}. 
In this paper, we introduce a robust, efficient, and interpretable methodology based on nonlinear manifold learning to detect anomalies in unsupervised settings where anomaly rate ($a_r$) is unknown and there is no training data with labeled anomalous samples\footnote{Our approach can naturally handle supervised and semi-supervised cases.}.

As a learning task, anomaly detection may be supervised, semi-supervised, or unsupervised \cite{mehrotra2017anomaly, noto2012frac,xia2022gan}. Supervised anomaly detection involves training a model on a labeled dataset where the normal and anomalous samples are known and labeled as such. In this case, the model learns to distinguish between normal and anomalous samples based on the features and patterns present in the training data and after training it is used to detect anomalies in  unseen data \cite{gornitz2013toward,pang2021toward}. Semi-supervised techniques require training a model on a dataset that consists of only normal samples and the trained model is then used to detect anomalous samples in unseen data \cite{ruff2019deep, villa2021semi,liu2021semi,de2021semi}. Unsupervised anomaly detection involves training a model on a dataset that includes normal and anomalous samples, with the goal of identifying samples that are unusual or do not conform to the expected pattern. In unsupervised anomaly detection, there is no pre-existing information that indicates which instances are considered normal. Moreover, the dataset is not divided into distinct training and testing phases as in supervised and semi-supervised settings \cite{chen2020unsupervised,cui2022survey,fraser2022challenges}.


As reviewed in \Cref{sec: background}, many different anomaly detection techniques have been recently developed in the literature \cite{usmani2022review,yang2022learning,alimohammadi2022performance,ergen2019unsupervised,ergen2019unsupervised}. However, these methods mostly require human supervision and their performance is very sensitive to $a_r$ (i.e., the number of anomalous samples in the data) \cite{fan2020robust}. This sensitivity is due to the fact that anomalies are rare events and their underlying distribution (which differs from the rest of the data) should be characterized based on very few samples. The performance of existing techniques is also very sensitive to the data dimensionality and their accuracy decreases as in high dimensions \cite{talagala2021anomaly}. 

\begin{figure*}[!b] 
    \centering
    \includegraphics[page=1, width = 1\textwidth]{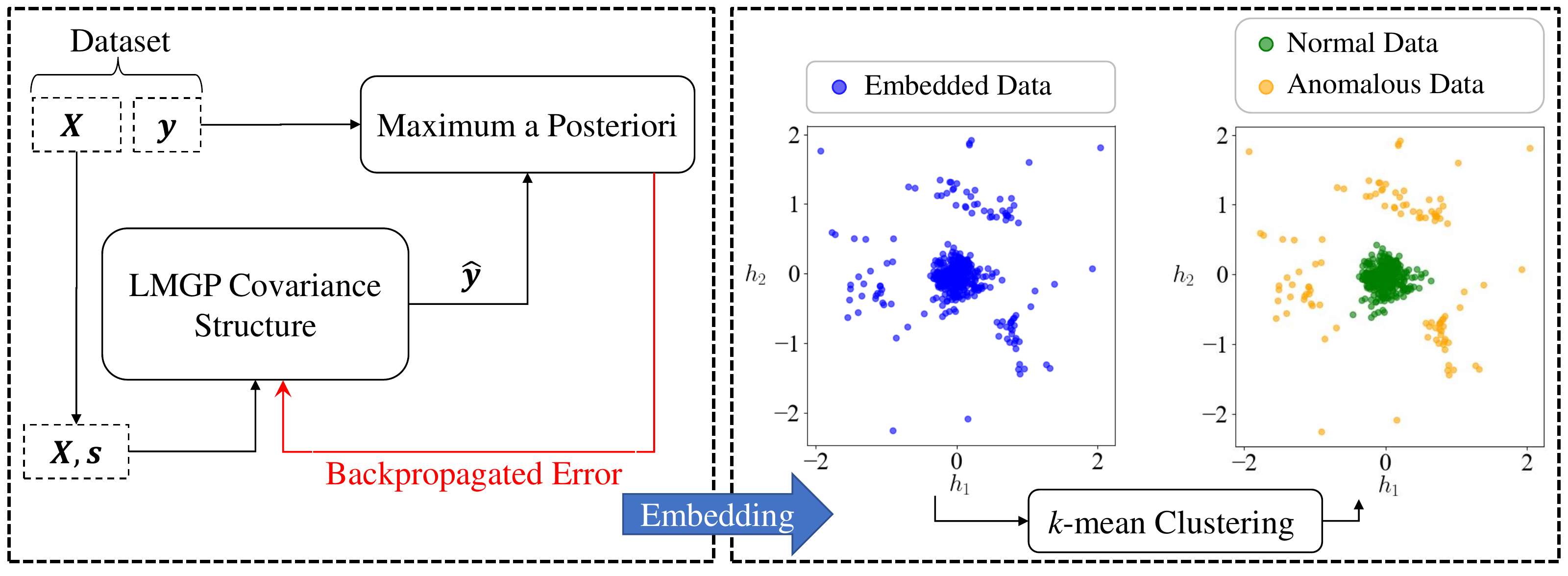}
    \vspace{-.5 cm}
    \caption{\textbf{Anomaly detection via manifold learning:} We use latent map Gaussian processes (LMGPs) to nonlinearly learn an embedding for each sample (this process relies on augmenting the features with the categorical variable $s$ which has as many unique levels as there are samples). We then use \kmean algorithm to cluster the learnt embedding into two sets where the samples in the larger set correspond to the normal data.}
    \label{fig: LMGP_block}
\end{figure*}

To address these challenges, in this paper we develop a robust anomaly detection technique based on nonlinear \manifoldlearning. 
Our approach leverages spatial random processes to, while offering a probabilistic perspective, learn a low-dimensional and visualizable manifold where anomalies are encoded with points that are distant from the rest of the data, see \Cref{fig: LMGP_block}. 
Our approach offers several major advantages compared to existing unsupervised anomaly detection methods. Firstly, it is non-parametric and does not require any parameter tuning. Secondly, it does not require any labeled examples or additional information such as the $a_r$. Lastly, it is well-suited for industrial applications that involve high-dimensional datasets with a limited number of samples. 

The rest of our paper is organized as follows. We review the major works on unsupervised anomaly detection in \Cref{sec: background}. Then, we introduce our approach in \Cref{sec: Proposed Approach} and test its performance against state-of-the-art techniques in \Cref{sec: results}. We conclude the paper with some final remarks and potential future directions in \Cref{sec: conclusion}.

    \section{Related Works} \label{sec: background}

In this section, we examine the key advancements and limitations of existing unsupervised anomaly detection methods which we classify into four main categories: neighbor-based, statistical-based, neural network (NN)-based, and hybrid methods that combine multiple techniques.

\subsection{Neighbor-based Methods} \label{sec: neighbor-based}

Neighbor-based anomaly detection methods utilize spatial information to differentiate between normal and anomalous data points. They compare a sample to its neighboring points and flag it as anomalous if it deviates significantly from its neighbors. Some neighbor-based methods consider the distance to the majority of data (i.e., they are distance-based) while other leverage the density of data surrounding a specific point (i.e., they are density-based \cite{breunig2000lof}).

Anomaly detection techniques require two components: an anomaly score and a threshold for detection. The anomaly score of a sample is a numerical value that indicates that sample's level of abnormality or deviation from expected behavior. The threshold is essentially a reference point for classifying data points as either normal or anomalous based on their respective anomaly scores. That is, by comparing the individual scores to the threshold, data points can be appropriately flagged as normal or anomalous. In distance-based anomaly detection methods, the criteria used for determining the anomaly score and threshold typically include the distance between a data point and its closest cluster center or the number of clusters surrounding the data point within a fixed distance \cite{pu2020hybrid, gao2012unsupervised, syarif2012unsupervised}. 
Similarly, in density-based methods such as the local outlier factor algorithm \cite{breunig2000lof}, the local density of a data point is compared to the local densities of its neighbors and the samples that exhibit a significantly lower density than their neighbors are flagged as anomalous.

Recently, distance-based methods are improved by leveraging low-rank and sparse matrix decomposition-distance \cite{zhang2015low} or Mahalanobis-distance-based classification \cite{magyar2022spatial} for both supervised and unsupervised anomaly detection. Another recent method is isolation forest which is a distance-based anomaly detection technique \cite{hariri2019extended}. Isolation forest is based on random forests and aims to find samples that are distant from the majority of the data. Isolation forest splits the data space using binary trees and assigns higher anomaly scores to data points that require fewer divisions to be isolated \cite{ song2021spectral, karczmarek2021fuzzy}. 

The power of neighbor-based algorithms lies in their unsupervised structure, i.e., they do not need a training dataset whose samples are labeled as either normal or anomalous. However, they are limited by their quadratic computational costs and potential inaccuracies when dealing with high-dimensional data. Alternative approaches or the integration of multiple methods are necessary to overcome the limitations associated with high-dimensional data. We elaborate more on this point in \Cref{sec: Hybrid Methods}.

\subsection{Statistical Methods} \label{sec: Statistica-based methods}

Statistical methods build probabilistic models on the data and leverage likelihood ratios to detect anomalies. That is, these methods presume an underlying probability distribution for the data and flag a sample as abnormal if its likelihood falls below 
a specified threshold \cite{wang2019outlier, rajabzadeh2016dynamic}. Determining this threshold is a difficult task in most applications. Moreover, these methods often leverage Gaussian processes (GPs) for constructing the likelihood which, while effective in semi-supervised settings \cite{lv2021latent, wang2019outlier}, face some challenges in unsupervised settings as determining the appropriate anomaly rates and thresholds are not straightforward. 
An example effort for addressing these issues is establishing the decision threshold based on the desired quantile of the approximating cumulative distribution function (CDF) \cite{yu2019unsupervised}. For the most part, however, these extensions do not scale well to high dimensions.

\subsection{Neural Network-based Methods} \label{sec: deep-based methods}

Deep learning (DL) is increasingly used for fault and anomaly detection \cite{pang2021deep,chalapathy2019deep}. Among the various DL models, autoencoders (AEs) have a particularly long history in industrial \cite{tao2022deep} as well as medical \cite{fernando2021deep, baur2021autoencoders} applications where typically convolutional auto-encoders \cite{hu2022video} and variational auto-encoders \cite {kingma2013auto} are used for anomaly detection. However, these methods are primarily trained to extract useful features from normal data and hence are best suited for supervised and semi-supervised anomaly detection scenarios. For instance, in \cite{lee2020unsupervised} an anomaly detection approach via convolutional AEs is proposed for a gas turbine operation. While this method is claimed to be unsupervised, its training phase relies on normal data which contradicts the definition of unsupervised anomaly learning. This issue is commonly seen in many works that leverage DL for anomaly detection.

\subsection{Hybrid Methods} \label{sec: Hybrid Methods}
Relying solely on a single algorithm may not produce satisfactory outcomes and hence hybrid methods use an ensemble of techniques to more accurately identify anomalies \cite{agrawal2015survey, zhang2023hybrid,yan2023hybrid}. Hybrid methods choose the ensemble members based on the application at hand and aim to leverage the strengths of each technique in handling particular anomaly detection cases. 
These approaches are particularly useful when dealing with high-dimensional data.
Some of these methods involve two separate steps \cite{aytekin2018clustering,} while others integrate the steps into a joint process \cite{ghafoori2020deep,zong2018deep}. 
For instance, \kmeans clustering is employed in \cite{aytekin2018clustering} to detect anomalies by grouping the normalization encoded values learnt via a deep AE. This work, however, does not leverage the reconstructions error (which is naturally provided by the AE) for anomaly detection which reduces its robustness. As part of our work, we extend this approach by redesigning the AE architecture based on our manifold learning technique (see \Cref{sec: LMGP}). Our extension achieves higher robustness and accuracy levels and we leverage it in our comparative studies in \Cref{sec: results}. 

In \cite{zong2018deep}, DAGMM is proposed which utilizes a deep AE to generate a low-dimensional representation for each data point by minimizing the reconstruction error. This representation along with the reconstruction error are then fed into a Gaussian Mixture Model (GMM) to detect anomalies. An estimation network is then used to facilitate the parameter learning of the mixture model. The DAGMM jointly optimizes the parameters of the deep AE and the mixture model's estimation network in an end-to-end fashion. However, this technique has some limitations that hinder its application in unsupervised anomaly detection. First, it requires knowledge of $a_r$. Second, it either needs to be trained on a dataset with a few anomalies or trained on a dataset containing only normal data which is contrary to the concept of unsupervised anomaly detection.

    \section{Proposed Approach} \label{sec: Proposed Approach}

Our anomaly detection approach has two primary stages. First, we use a nonlinear \manifoldlearning algorithm to map all the data points into a low-dimensional and visualizable space that preserves the underlying structure of the data while separating normal samples from the anomalous ones. This manifold assigns a latent point to each sample and aims to learn patterns and dependencies that are typically too difficult to discern in the original feature space (due to high dimensionality or complexity). After the manifold is built, we apply a clustering algorithm to it to group the encoded latent points into clusters based on their positions in the manifold, i.e., our similarity metric is automatically learnt since it is the distances between the latent points. This clustering enables us to distinguish between anomalous and normal samples and identify potential outliers that deviate significantly from the expected behavior. In particular, we always cluster the latent points into two groups and label the larger group as normal. Our clustering and labeling decisions are based on the facts that $(1)$ a sample is either normal or anomalous, and $(2)$ anomalies are, by nature, rare events and hence the data should contain far more normal data than anomalies.

In our approach we leverage the fact that the normal data are all generated via the same underlying source while anomalous samples either are generated by different sources, or are normal samples that are corrupted during the data collection/recording process. Besides making this distinction on the distributional characteristics, we make no assumptions about the underlying nature of distributions (e.g., anomalies can be generated by multiple distinct sources, see \Cref{sec: results} for an example). In essence, our manifold is expected to solve the following inverse problem: distinguish between the sources of normal and anomalous data given a set of unlabeled samples where all the normal points are generated by a single source. We propose to build these manifolds via either latent map Gaussian processes (LMGPs, which are our preferred approach) or AEs whose architecture are inspired via LMGPs.

In what follows, we first introduce LMGPs and then deep AEs for manifold learning. Then, in \Cref{sec: \kmean} we provide a detailed explanation of how we use \kmeans clustering to identify anomalous points on the learned manifolds. Lastly, we delineate the advantages of our proposed approaches over existing techniques in the literature. 

\subsection{Nonlinear \manifoldlearning with Latent Map Gaussian Processes} \label{sec: LMGP}
GPs are widely used metamodels that assume the training data is generated from a multivariate normal distribution with parametric mean and covariance functions. These assumptions enable the use of closed-form formulas based on the conditional distributions to predict unseen events. 
Conventional GPs do not accommodate categorical or qualitative features as these variables are not equipped with a distance measure. LMGPs are extensions of GPs that address this limitation \cite{oune2021latent} by automatically learning an appropriate distance metric for categorical variables. In this paper, we use this learning ability of LMGPs to distinguish between normal and anomalous samples. To this end, we augment the input space with the qualitative feature $s$ whose number of unique levels (i.e., distinct categories) matches with the number of samples, i.e., $s=\{ '1', \cdots, 'n'\}$ where $n$ is the dataset size. After this augmentation, we train an LMGP as described below and then use the manifold that it learns for $s$ in \Cref{sec: \kmean} for anomaly detection.

We presume the general case where the original dataset has $dx$ numerical and $dt$ categorical inputs which are denoted via $\boldsymbol{x}=[x_1, \ldots, x_{dx}]^T$ and $\boldsymbol{t}=[t_1, \ldots, t_{dt}]^T$, respectively. With the addition of $s$, the mixed input space is given by $\boldsymbol{u}=[x_1, ..., x_{dx}, t_1, ..., t_{dt},s]^T$. Given the $n$ training pairs $(\ub^i, y^i)$ where $y$ is the response, LMGP assumes that the following relation holds:

\begin{equation} 
    \begin{split}
        y(\boldsymbol{u})={\beta}+\xi(\boldsymbol{u})
    \end{split}
    \label{eq: GP-prior}
\end{equation}

\noindent where ${\beta}$ is an unknown constant and $\xi(\boldsymbol{u})$ is a zero-mean GP with the covariance function or kernel of:

\begin{equation} 
    \begin{split}
        \operatorname{cov}\left(\xi(\boldsymbol{u}), \xi\left(\boldsymbol{u}^{\prime}\right)\right)=c\left(\boldsymbol{u}, \boldsymbol{u}^{\prime}\right)=\sigma^2 r\left(\boldsymbol{u}, \boldsymbol{u}^{\prime}\right)
    \end{split}
    \label{eq: GP-Cov1}
\end{equation}
\noindent where $\sigma^{2}$ indicates the variance of the process and $r(.,.)$ is the parametric correlation function. Evaluation of $r(.,.)$ in \Cref{eq: GP-Cov1} requires the conversion of all categorical variables (i.e., $\tb$ and $s$) to some numerical features which are embedded in one or multiple manifolds\footnote{Regardless of the number of manifolds, we always use two dimensional manifolds which have large learning capacity while being very easy to interpret and visualize.}. Here, we consider two separate quantitative manifolds where the first one encodes $\tb$ (i.e., each unique combination of $\tb$ is encoded with a single point in this manifold) while the other one encodes $s$ (i.e., each data point or level of $s$ is mapped to a single point in the this manifold). We make this decision to increase the interpretability and accuracy of our approach since visualizing and clustering the latent points corresponding to the levels of $s$ is much easier and more robust in a 2D space which does not encode $\tb$.

For an LMGP with two manifolds, we propose the following Gaussian correlation function that is tailored for anomaly detection:
\begin{align}
    r(\begin{bmatrix}
        \boldsymbol{x}\\
        \boldsymbol{t}\\
        s
    \end{bmatrix} , 
    \begin{bmatrix}
        \boldsymbol{x^\prime}\\
        \boldsymbol{t^\prime}\\
        s^{\prime}
    \end{bmatrix}) = 
    \exp \{-\sum_{i=1}^{dx} 10^{\omega_i}(x_i-x_i^{\prime})^2\ -
    \sum_{i=1}^{dz}(z_i(\boldsymbol{t})-z_i(\boldsymbol{t}^{\prime}))^2 -
    \sum_{i=1}^{dh}(h_i(s)-h_i(s^\prime))^2\}
    \label{eq: LMGP-Corelation-extended2}
\end{align}

\noindent where $\boldsymbol{\omega}=[\omega_{1}, \ldots, \omega_{dx}]^{T}$ are the scale parameters associated with the numerical features $\xb$, $\zb$ is a $dz$ dimensional vector that encodes $\tb$, and $\hb$ is a $dh$ dimensional vector that encodes $s$. 
To learn the variables in the $z-$sapce, LMGP first assigns a unique numerical vector (i.e., a prior representation) to each combination of categorical variables and then maps these vectors to the $z-$ space (where the posteriors reside) via a parametric function (a similar process is done to learn the latent variables in the $h-$ space).
Here, we use grouped one-hot encoding for designing the prior vectors and leverage a linear transformation to map them into the respective manifolds. That is:
\begin{subequations} 
\begin{equation} 
    \begin{split}
     \boldsymbol{z(t)}=\boldsymbol{\zeta}_1(\tb) \boldsymbol{A}_z,
    \end{split}
    \label{eq: LMGP-zeta2}
\end{equation}
\begin{equation} 
    \begin{split}
        \boldsymbol{h}(s)=\boldsymbol{\zeta}_2(s) \boldsymbol{A}_h
    \end{split}
    \label{eq: LMGP-zeta2}
\end{equation}
\end{subequations} 

\noindent where the rectangular matrix $\boldsymbol{A}_z$ maps $\boldsymbol{\zeta}_1(\tb)$ to $\boldsymbol{z(t)}$ which are, respectively, the prior and posterior numerical representations of $\boldsymbol{t}$. Similarly, $\boldsymbol{A}_h$ maps $\boldsymbol{\zeta}_2(s)$ to $\boldsymbol{h}(s)$ which denote, respectively, the prior and posterior representations of $s$.

The hyperparameters of an LMGP include  $\beta, \boldsymbol{A}_z, \boldsymbol{A}_h, \boldsymbol{\omega}$, and $\sigma^{2}$ which we collectively denote via $\Thetab$ and jointly estimate them via maximum a posteriori (MAP):
{\begin{equation} 
    \begin{split}
        \widehat{{\Thetab}} = 
        \underset{{\Thetab}}{\operatorname{argmax}} \hspace{3mm} |2 \pi \sigma^2 \boldsymbol{R}|^{-\frac{1}{2}} \times \exp \left\{ \frac{-1}{2}(\boldsymbol{y}-\boldsymbol{1} {\beta})^T(\sigma^2 \boldsymbol{R})^{-1}(\boldsymbol{y}-\boldsymbol{1} {\beta}) \right\} \times 
        \mathrm{P}(\cdot)
    \end{split}
    \label{eq: MAP-seapare-A}
\end{equation}}

\noindent or equivalently:
\begin{equation} 
    \begin{split}
        \widehat{{\Thetab}} =
        \underset{{\Thetab}}{\operatorname{argmin}} \hspace{3mm} \frac{n}{2} \log \left(\sigma^2\right)+\frac{1}{2} \log (|\boldsymbol{R}|)+\frac{1}{2 \sigma^2}(\boldsymbol{y}-\boldsymbol{1} {\beta})^T \boldsymbol{R}^{-1}(\boldsymbol{y}-\boldsymbol{1} {\beta}) 
        +\log(\operatorname{P(\cdot)}) 
    \end{split}
    \label{eq: LMGP-parameter2}
\end{equation}

\noindent where $\log (\cdot)$ and $|\cdot|$  stand for the natural logarithm and the determinant operator, respectively. Also, $\boldsymbol{y}=[y^1, \ldots, y^n]^T$ denotes the $n \times 1$ vector of outputs in the training data, $\boldsymbol{R}$ is the $n \times n$  correlation matrix with the $(i, j)^{t h} \text { element } R_{i j}=r(\boldsymbol{u}^i, \boldsymbol{u}^j) \text { for } i, j=1, \ldots, n$. Also, $\boldsymbol 1$ is the $n \times 1$ vector of 1 and $P(\cdot)$ indicates the prior on the hyperparameters. Herein, we follow \cite{Gpytorch} and place independent priors on the hyperparameters in which $\sigma\sim Lognormal(0,3)$, while $\boldsymbol \omega, \beta, \boldsymbol{A}_z$ and $\boldsymbol{A}_h$ follow normal priors as {$\boldsymbol\omega \sim \mathcal N(-3,3),\beta \sim \mathcal N(0,1), \boldsymbol{A}_z \sim \mathcal N(0,3)$ and $\boldsymbol{A}_h \sim \mathcal N(0,3)$. 

The objective function in Eq. \ref{eq: LMGP-parameter2} can be effectively and quickly minimized via gradient-based optimization techniques \cite{RN783, RN649}. The formulations mentioned above can be adapted to handle datasets with noise by introducing a nugget or jitter parameter, $\delta$ \cite{bostanabad2018leveraging}. As a result, $\boldsymbol{R}$ is replaced by $\boldsymbol{R}_{\delta}= \boldsymbol{R} + \delta I_{n\times n}$, where $I_{n\times n}$ is a $n \times n$ identity matrix. 

We now elaborate on our rationale for identifying anomalies based on the latent points that encode the levels of the categorical variable $s$: 
Factors that render a sample anomalous are potentially many. Moreover, these factors are typically of different nature and we may not even know all of them (i.e., unknown sources cause anomalies). Consider a manufacturing process where the performance metrics of the built sample can be affected via human errors, random variations in the raw materials, faulty measurements, or process variations. Since it is extremely difficult to characterize all these factors for all the samples and then use them for detecting anomalies, we assign a unique feature to each sample. This features must be categorical since a numerical one (that is directly assigned by the analyst) induces incorrect relations among the samples. To learn the relation between the levels of this categorical feature (or, equivalently, learn the relation among the samples), we build an LMGP model as described above. In particular, LMGP maps the levels of $s$ to points in the $h-$ space such that they optimize \Cref{eq: LMGP-parameter2}. 

The optimization in \Cref{eq: LMGP-parameter2} requires normal points to be encoded via close-by points which are all distant from anomalies (note that latent points that encode anomalies can be distant from one another in the $h-$ space depending on whether they are affected by the same factors). 
We explain this requirement via \Cref{eq: LMGP-Corelation-extended2} for the case where $\xb \approx \xb ^\prime$, $\tb \approx \tb ^\prime$, and $y \approx y ^\prime$ where the first two conditions imply $r(\ub, \ub^\prime) = exp(- \sum_{i=1}^{dh}(h_i(s)-h_i(s^\prime))^2)$. That is, for two samples that are very close in the input space and have similar response values, the correlation function is maximized (and hence the optimization problem in \Cref{eq: LMGP-parameter2} is minimized) if those two points are encoded with close-by points in the $h-$ space, i.e., $\Vert \hb(s) - \hb(s^\prime) \Vert^2 << 1$.

\subsection{Nonlinear \Manifoldlearning with Autoencoders} \label{sec: AE}

The second method that we develop for manifold learning is based on AEs which are widely used for anomaly detection. 
Most existing approaches that employ AEs for anomaly detection assume that the trained AE cannot accurately reconstruct anomalies as the underlying distribution that the AE learns does not capture the anomalies well.
While this scenario might be the case for some anomalous samples, there are cases where we observe anomalies with relatively small reconstruction errors. To showcase such cases, we consider the following two cases:
$(1)$ if the anomalies are generated by distributions that are easy to learn, the AE can effectively characterize them and achieve low reconstruction errors even for anomalies, and 
$(2)$ if normal samples are noisy or display complex structures, the AE may achieve relatively large reconstruction errors even for these samples which makes it difficult to distinguish them from the anomalies. 
Therefore, relying solely on the reconstruction errors achieved by an AE does not provide robust performance for anomaly detection. 

To address this robustness issue, in addition to using the reconstruction errors, one can leverage the learnt manifold of an AE which is expected to encode anomalies far from the normal samples. 
As reviewed in \Cref{sec: Hybrid Methods}, the DAGMM method \cite{zong2018deep} leverages this idea but has two major shortcomings (see \Cref{sec: Hybrid Methods} for details). 
To address these limitations in the context of regression (i.e., when the data has inputs and outputs), we draw inspiration from LMGPs. In particular, we argue that jointly using the total reconstruction error and the encoded positions in the manifold for anomaly detection via AEs can result into identifiability issues since the latter is learnt by minimizing the former while training the AE. Hence, we propose to use the latent points and part of the reconstruction error that characterizes the error on reproducing the responses. 

Our AE has the typical architecture that consists of the encoder $f(\cdot)$ and the decoder $g(\cdot)$. $f(\cdot)$ takes the inputs and response of the $i ^{th}$ sample and maps this data point to the low-dimensional $h-$ space. The encoded value of $i ^{th}$ sample is $h_1^i = f([\boldsymbol{x}^i, y^i])$ which is then fed into $g(\cdot)$ to reconstruct the $i^ {th}$ sample:
\begin{equation} 
    \begin{split}
[\widehat{\boldsymbol{x}}^i,\widehat{y}^i]=g \left ( h_1^i \right )  = g \left ( f \left ([\boldsymbol{x}^i, y^i]\right) \right )  
    \end{split}
    \label{eq: AE outpout}
\end{equation}
where $\widehat {\boldsymbol{x}}^i $ and $\widehat{y}^i$ are the reconstructed counterparts of $\boldsymbol{x}^i$  and $y^i$, respectively. We train our AE by minimizing the total reconstruction error on both the inputs and output. That is:
\begin{equation} 
    \begin{split}
        l = 
        \frac{1}{n} \sum_{i=1}^{n}  \left(  \left\| \boldsymbol{x}^i-\widehat {\boldsymbol{x}}^i \right\|_2^2  + (y^i-\widehat{y}^i)^2 \right)
    \end{split}
    \label{eq: Loss_autoencoder}
\end{equation}
\noindent where $\left\| \cdot \right\|$ denotes Euclidean norm operator and ${n}$ denotes the total number of samples in the dataset\footnote{If the dataset has categorical inputs ($\boldsymbol{t}$) one-hot encoding should be used before feeding data to the AE.}. The loss function in \Cref{eq: Loss_autoencoder} is the typical mean squared error (MSE) that is used in training AEs. 

Once the AE is trained, we create a secondary manifold whose axes are $h_1$ (which is the bottleneck of the AE and encodes the samples) and $h_2^i=|\widehat {y}^i-y^i|$ which measures the error of AE in reproducing the response values. We then use this secondary manifold for anomaly detection, see \Cref{fig:Autoencoder_block}. 

\begin{figure*}[!t] 
    \centering
    \includegraphics[page=1, width = 1\textwidth]{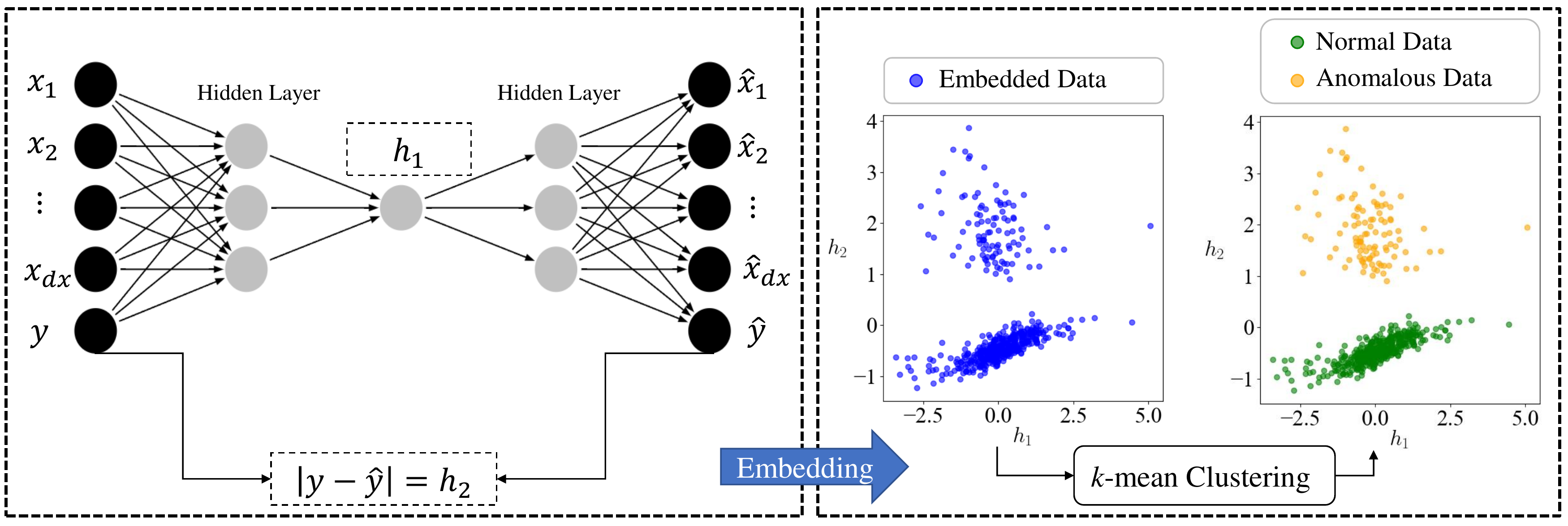}
    \vspace{-.5cm}
    \caption{\textbf{Anomaly detection with autoencoders:} We feed the input and output of the data to the deep AE which aims to reconstruct each sample while passing it through the one-dimensional $h-$ space. We then use the encoded value for each sample (i.e., $h_{1}$) along with the corresponding reconstruction error (i.e., $h_{2}=|\widehat{y}-y|$) to generate a $2D$ manifold. Finally, we use the \kmeans clustering algorithm to group similar data points and identify anomalies.}
    \label{fig:Autoencoder_block}
\end{figure*}

\subsection{Manifold-based Clustering for Anomaly Detection} \label{sec: \kmean}

With either LMGP or AE, we map each sample to a point in the $h-$space where similar data (i.e., samples that have the same underlying generative distribution) are expected to be close-by and distant from the anomalies. Since anomalies are rare, we expect the majority of the samples to be encoded with points that are close to the origin of the manifold (i.e., $\Vert \hb_i \Vert^2 << 1$ for the $i^{th}$ sample) since we are using MAP and otherwise the loss function of the  LMGP is not optimally minimized. We test this expectation in \Cref{sec: results} where we show that LMGP consistently outperforms AE in this regard.

Regardless of where the samples are encoded in the $h-$space, we cluster them into two groups (as we know the data points are either normal or not) and label the samples that belong to the larger cluster as normal (since anomalies are rare events). 
To this end, we employ the \kmean clustering algorithm which groups the points via an iterative scheme where it first assigns each data point to the cluster whose centroid in the $h-$space is closest to that data point. It then updates the cluster centroids to be the mean of all the points assigned to that cluster. This assignment-update process is repeated until the centroids no longer change or a maximum number of iterations is reached. 
In our anomaly detection framework, the input to the algorithm is the distance of each point to the center of the learned manifold and the number of clusters $k=2$ while the output is a labeled manifold where each point is assigned to one of the clusters, see \Cref{fig: LMGP_block} and \Cref{fig:Autoencoder_block}. 
With LMGP, we first augment the input space with the qualitative feature $s$ whose number of unique levels (i.e., distinct categories) matches with the number of samples. After LMGP is fit, we feed the manifold that it learns for $s$ to the \kmeans clustering algorithm to identify the anomalies. With AE, we first train it to minimize the total reconstruction error in \Cref{eq: Loss_autoencoder}. Then, we build the secondary manifold (based on $h_{1}$ and part of the reconstruction error which is $h_{2}=|\widehat{y}-y|$) and use it for clustering. Note that in AE-based approche the scale of these two components of the manifold can vary. Therefore, we utilize their normalized values to construct the manifold.

\subsection{Comparison to Existing Techniques} \label{sec: Advantagess of Proposed Approachs}

Our anomaly detection method is a hybrid one in essence since we first learn a low-dimensional representation of the data and then apply a clustering algorithm to it. The major difference between our approach and the existing technologies, is its first stage. Our LMGP-based technique is unique since it uses an extension of GPs for \manifoldlearning in the context of anomaly detection. The proposed technique offers several advantages compared to other methods. Firstly, due to the non-parametric nature of GP, the proposed technique does not require any parameter tuning, which is a significant challenge in deep AE-based methods. Secondly, our method does not require any labeled examples or additional information such as $a_r$ which are typically needed in other unsupervised approaches. Thirdly, our proposed technique is well-suited for industrial applications that involve high-dimensional datasets with a limited number of samples. Fourthly, our LMGP-based anomaly detection has a significant advantage over other anomaly detection methods as it automatically handles mixed input spaces (that have both categorical and numerical features) which ubiquitously arise in design applications such as material composition optimization.

Inspired by our LMGP-based anomaly detection method, we introduce the AE-based approach for anomaly detection which works based on nonlinear manifold learning. Our  AE-based method tackles two critical challenges encountered by most existing AE-based techniques in the literature. Firstly, it does not require knowledge of $a_r$ which is often unknown in realistic applications. Secondly, it operates in a completely unsupervised manner as it does not involve a training phase with only normal data.

    \section{Experiments} \label{sec: results}
In this section, we test the performance of our approach against isolation forest and DAGMM which are two popular anomaly detection methods. 
We consider three analytic and two real-world examples in \Cref{sec: Analytic Examples} and \Cref{sec: Real-world Datasets}, respectively. In each experiment, we repeat the simulation $20$ times to assess the robustness and consistency of the results. 

We follow two mechanisms for generating anomalous samples. 
In the first one we generate anomalous samples via one or multiple functions which are different than the function that generates the normal data. This mechanism aims to model scenarios where systematic or malicious errors corrupt the data.
In the second mechanism we select a random subset of the data and corrupt the outputs via the following equation:
\begin{equation} 
    \begin{split}        
     {y_{a}(\boldsymbol{x})=(1+ a ) \times y_{n}(\boldsymbol{x})}
    \end{split}
    \label{eq: anomaly generator}
\end{equation}
where $y_{a}$ and $y_{n}$ indicate anomalous and normal outputs, respectively. Also, parameter $a$ represents the anomaly bound and is randomly selected for each sample from a uniform distribution on interval $[1, 2]$. This mechanism serves to assess the effectiveness of our methods in scenarios such as temporary measurement errors and fluctuating environmental conditions.

In the current study, the anomalous and normal samples are categorized as the positive and negative classes, respectively. Throughout this section, we use ${F_1}$-score and G-mean metrics to evaluate the performance (see \Cref{sec: Metrics definition} for definitions). We also provide the performance of all approaches in terms of precision in \Cref{sec: Precision}. These metrics quantify the accuracy of each method in correctly classifying both normal and anomalous samples. 

\subsection{Analytic Datasets} \label{sec: Analytic Examples}
We conduct three experiments where the data are generated by the analytic functions detailed in \Cref{sec: appendix-equation}. The dimensionality of the output/response space is one in all of these cases while the input space is either 10 or 8 dimensional. For each analytic example we generate a total of $500$ samples and control the number of anomalous samples via $a_r$. To assess the sensitivity of each approach to the number of abnormal data, we consider the four anomaly rates of $[0.05, 0.1, 0.2, 0.3]$ (e.g., $0.05$ means that $0.05\times500$ of the samples are turned into anomalous data via either mechanism one or two). 
To challenge all the anomaly detection methods, we add noise to all the $500$ data points where the noise variance is defined based on the range of each function (see \Cref{sec: appendix-equation} for details).

\subsubsection{Multiple Anomalous Data Sources} \label{sec: Wing_Example_1}
We use variations of the Wing model \cite{moon2010design} that estimate the weight of a light aircraft wing (see Appendix \ref{sec: appendix-equation} for the functional forms). Specifically, we utilize three different sources modeled via \Cref{eq: Wing_HF-function} through \Cref{eq: Wing_LF2-function} where Source $1$ generates normal data while the other two sources generate anomalous samples. The total number of samples is $500$ and for each value of $a_r$ we equally sample from sources $2$ and $3$ (e.g., $a_r=0.1$ corresponds to a case where $450$, $25$, and $25$ samples are generated via sources $1, 2,$ and $3$, respectively). 

\begin{figure}[!t]
    \centering
        \includegraphics[width=1\linewidth]{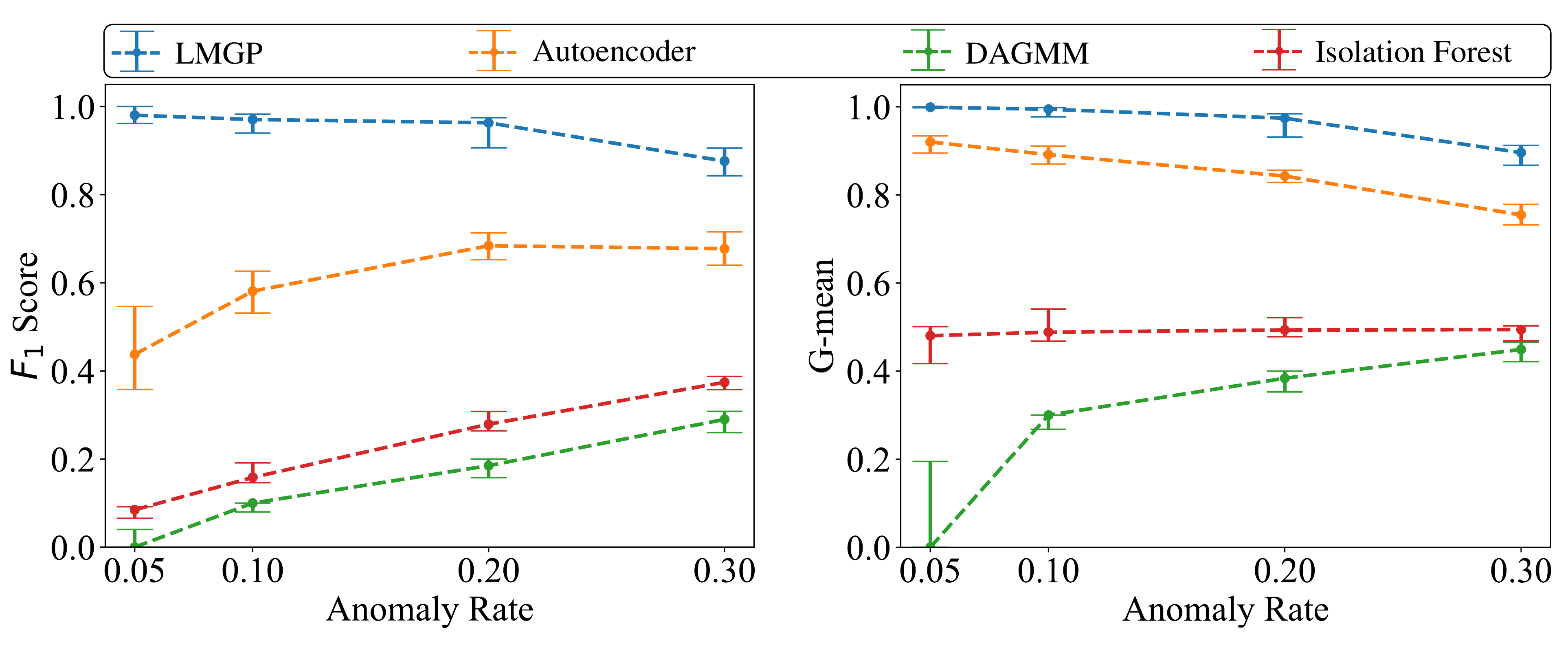}
    \vspace{-.7cm}
    \caption{\textbf{Effect of anomaly rate on the performance of different approaches:} 
    In all cases, our approach that is based on either LMGP or AE outperforms DAGMM and isolation forest which require the knowledge of $a_r$. As $a_r$ increases, the $F_1-$ score of all approaches increases except for LMGP (see the main text for the reasons behind this trend). Additionally, all methods achieve similar results across the $20$ repetitions as hence the error bars are pretty narrow. 
    }
    \label{fig: Wing_results_case_1}
\end{figure}
Figure \ref{fig: Wing_results_case_1} depicts the results of the LMGP-based, AE-based, DAGMM, and isolation forest methods in this example as $a_r$ increases from $0.05$ to $0.30$.  As the trends indicate, our LMGP-based approach consistently outperforms other approaches and is followed by our AE-based anomaly detection method. We attribute the superior performance of LMGP to its probabilistic nature and the fact that its learnt manifold effectively separates the anomalies from the rest of the data by optimizing \Cref{eq: LMGP-parameter2}. Such a separation makes clustering easier as well. 
However, our AE approach builds its manifold based on (part of) the reconstruction error and the encoding (see \Cref{fig:Autoencoder_block}) which are sensitive to variations and noise in small datasets and relatively high dimensions. 

We observe in \Cref{fig: Wing_results_case_1} that as $a_r$ increases the $F_1$ score of all the approaches increases except for LMGP. This exception is due to the fact that the total number of samples is always fixed to $500$ and increasing $a_r$ means that the dataset has increasingly more anomalous samples (and less normal data) which forces LMGP to learn a more complex distribution that can explain the behavior of both normal and anomalous data (as opposed to only learning the underlying distribution of the normal samples). For a similar reason, the G-mean score of our LMGP-based approach decreases as $a_r$ increases.

In the case of DAGMM (which directly uses $a_r$ as one of its parameters), it cannot detect anomalies at low $a_r$ since $(i)$ it has access to insufficient information on anomalies, and $(ii)$ our datasets are noisy. Specifically, at $a_r=0.05$ the model tends to randomly label approximately $5$ percent of the data (mostly normal ones) as anomalies and the true positive rate is observed to be close to zero. The findings from \cite{zong2018deep} support our observation and demonstrate that the performance of DAGMM substantially decreases as the anomaly rate decreases (see the Thyroid example discussed in \cite{zong2018deep}).

In the case of the isolation forest our results indicates that it divides the dataset into two classes with approximately the same size regardless of the $a_r$ value. That is, as the number of anomalies increases the true positive rate (correctly identified anomalies) and false negative rate (incorrectly classified normal samples) both increase while decreasing the true negative rate and the false positive rate (normal samples falsely identified as anomalies). Consequently, the G-mean value (which provides an overall performance measure) does not change because the increase in true positives is offset by the decrease in true negatives. 
However, these adjustments lead to an improvement in the $F_1$ score as its numerator grows more rapidly than its denominator (note that, as shown in \Cref{sec: Metrics definition}, the true positive rate is the only term in the numerator and directly affects the $F_1$ score while in the denominator there are three terms where two of them cancel each other out (the increase in false negative and decrease in false positive counterbalance each other).

\begin{figure}[!b]
    \centering
    \includegraphics[width=1\linewidth]{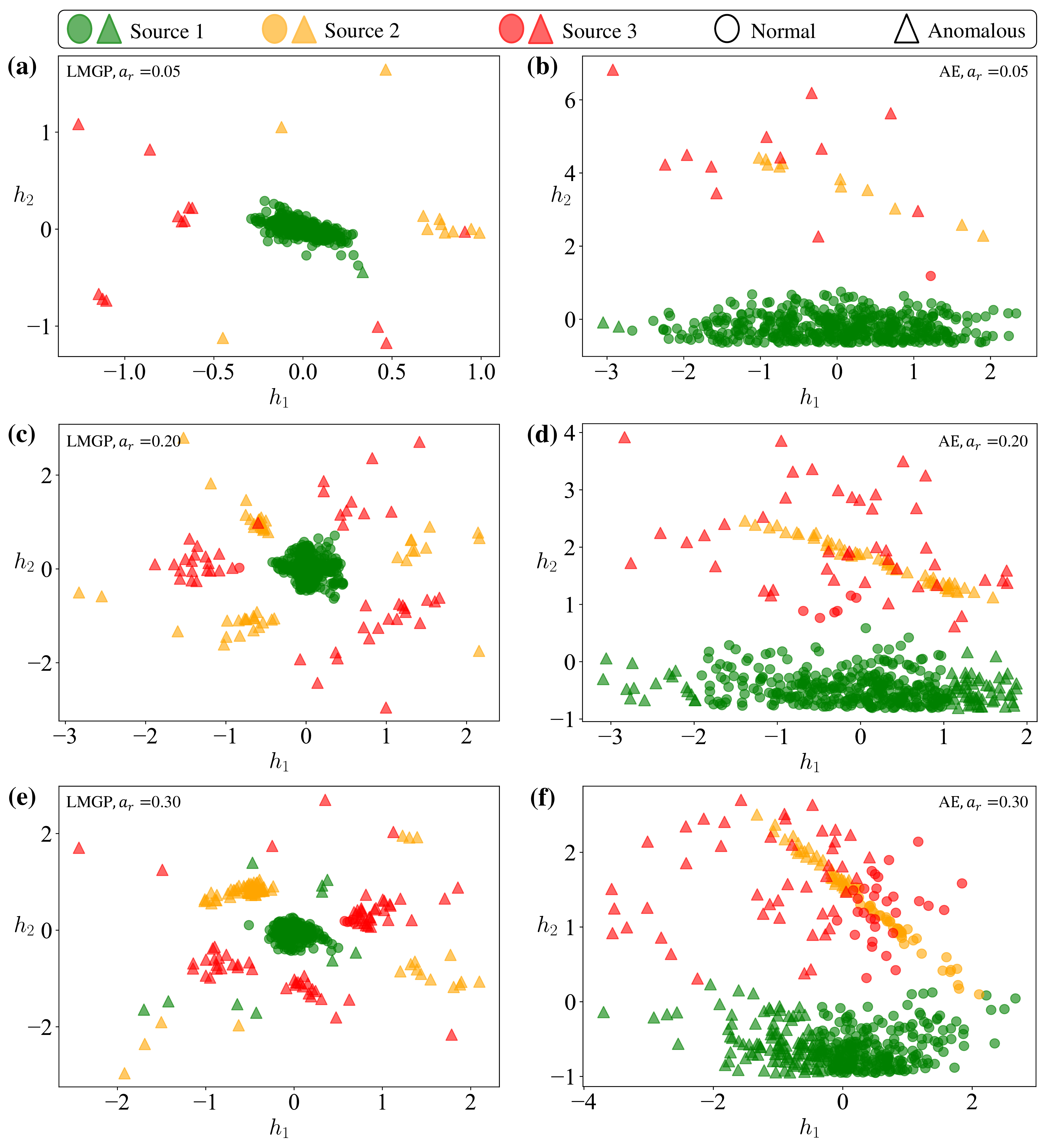}
    \vspace{-.5cm}
    \caption{\textbf{Learned manifold in Wing example with multiple anomalous data sources:} We illustrate the learned manifolds using our two different techniques (LMGP and deep AE) and for three different anomaly rates. The shape assigned to each point on the learned manifold indicates the result of our anomaly detection process (circles and triangles represent samples identified as normal and anomalous, respectively). The colors reflect the ground truth information (not used in our approach) where green, orange, and red correspond to samples from Source $1$, Source $2$, and Source $3$, respectively. Comparing the two plots in each row we observe that the manifold of LMGP is more interpretable and also results into a more accurate identification of normal and anomalous samples. By increasing $a_r$, the performance of LMGP drops while that of the AE increases. 
    }
    \label{fig: Wing_manifold_case_1}
\end{figure}

In the case of our AE-based method, a similar situation happens where its $F_1$ score improves as more anomalous data are provided to it. That is, the model has access to more information on anomalies and hence can build a manifold that betters distinguishes between the normal and anomalous data. However, as $a_r$ increases, the information on normal data decreases (which prevents the AE from effectively encoding normal samples into its manifold) and hence its G-mean drops. 

We note that the performance drop of LMGP (in terms of either $F_1$ or G-mean) is expected since it operates under the assumption that anomalies are rare events. In our studies, the high anomaly rates in \Cref{fig: Wing_results_case_1} are meant to test the sensitivity of different approaches to this assumption in extreme cases which rarely happen in realistic applications. 

To gain more insights into the performance of our approaches, we examine their learnt manifolds. 
The learned manifolds for this example with three different anomaly rates are presented in \Cref{fig: Wing_manifold_case_1} for one of the randomly selected repetitions. In these plots the shape assigned to each latent point represents the result of our anomaly detection process (circles represent normal samples while triangles flag anomalous data). Additionally, the points on the learned manifold are color-coded based on the ground truth information where green, orange, and red correspond to samples from Source $1$, Source $2$, and Source $3$, respectively (note that the ground truth is not used by either of our approaches). This combination of shape and color enables easy interpretation of both the detected anomalies and the ground truth information within the manifolds. 
 
As depicted in \Cref{fig: Wing_manifold_case_1}, both approaches aim to distinguish between normal and anomalous points.  However, the spread of the latent points are quite different: while LMGP mostly places the normal data in a circle centered at the origin (which means that both latent variables are equally important for classification via \kmeans clustering), AE maps them to a relatively stretched area and primarily relies on $h_2$ for classification. While our use of $h_2$ dramatically improves the performance of DAGMM (see \Cref{sec: AE} and \Cref{fig: Wing_results_case_1}), it is still not as effective as LMGP. 

It is important to note that a portion of the misclassification errors associated with our methods is attributed to the \kmean algorithm rather than LMGP or AE. For instance, in \Cref{fig: Wing_manifold_case_1} \textbf{(a)}, there is an instance where a normal sample is misclassified (see the green triangle close to the origin). While LMGP effectively encodes this point close to the other normal samples, the \kmean algorithm misclassifies it as an anomaly due to its relatively greater distance from the center of the manifold compared to the other normal points. 

Lastly, we note that the latent space of neither of our approaches is robustly able to distinguish between the different anomaly sources. That is, we cannot specify the number of different mechanisms that result into anomalous behavior. However, this behavior is expected to some extent since our approaches are not designed to identify the number of different mechanisms that result into abnormal behavior.

\subsubsection{Wing Model with One Anomaly Source} \label{sec: Wing_Example_2}

\begin{figure}[!b]
    \centering
        \includegraphics[width=1\linewidth]{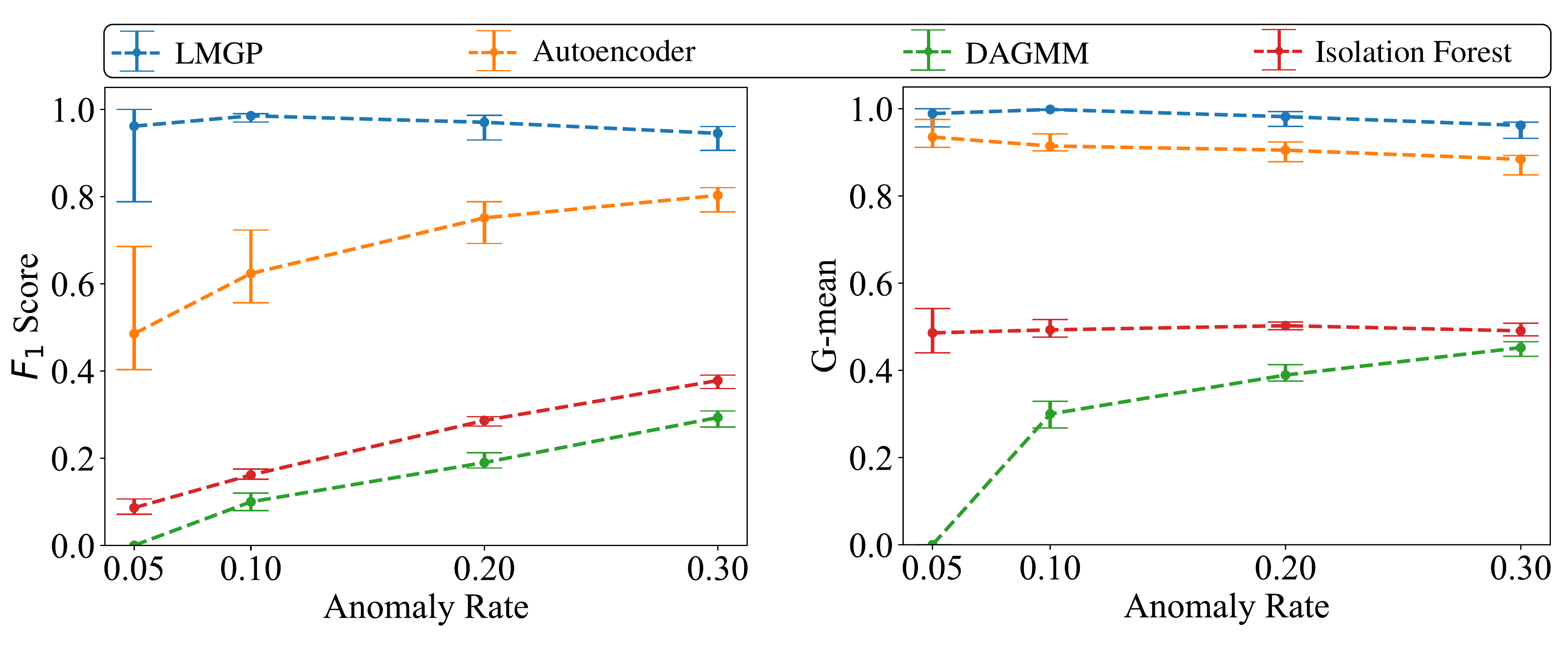}
            \vspace{-.7cm}
    \caption{\textbf{Anomaly detection results in the Wing model with one anomaly source:}  The results clearly indicate that our LMGP-based and deep AE-based methods outperform isolation forest and DAGMM. See the main text for the rationale behind the trends in both $F_1$ score and G-mean plots.}
    \label{fig: Wing_results_case_2}
\end{figure}
In this example, we generate $500$ samples from Source $1$ of the Wing model in \Cref{eq: Wing_HF-function} and then corrupt a subset of this dataset following \Cref{eq: anomaly generator} and for various values of $a_r$.
The comparison results are summarized in \Cref{fig: Wing_results_case_2} which clearly illustrates the consistent superiority of our LMGP-based technique over all other methods especially in correctly identifying the anomalies (see the $F_1$ score plot). The overall trends in this figure are quite similar to those in \Cref{fig: Wing_results_case_1} but there are two notable differences in the case of our LMGP-based approach. 
Firstly, both $F_1$ and G-mean scores slightly improve as $a_r$ is increased from $0.05$ to $0.10$ but they both drop as $a_r$ exceeds $0.10$.
Secondly, when $a_r=0.05$ the variation of $F_1$ score in LMGP's results is significantly higher compared to the previous example which indicates that the performance is more sensitive across the $20$ repetitions in this example.

We attribute these changes in LMGP's results to the stochastic nature of the anomalies. When $a_r$ is set to $0.05$ the model has only $25$ anomalous samples. In the previous example (\Cref{sec: Wing_Example_1}), these anomalies are highly correlated and come from specific sources. In contrast, the anomalies in this example are produced via a stochastic process with low correlation which produces a more diverse distribution of anomalies throughout the dataset. 
Consequently, LMGP cannot accurately identify that the underlying distribution of these anomalies is different than the rest of the data which, in turn, challenges the clustering stage as LMGP's manifold does not very accurately distinguish normal and anomalous samples. 
As $a_r$ is increased to $0.10$ and more anomalous samples are included in the data, LMGP can better learn the underlying mechanism that generates the anomalies and thus both $F_1$ and G-mean scores increase and the variations across the repetitions decrease.
However, this positive trend is reversed as $a_r$ exceeds $0.10$ since LMGP either cannot learn the normal data as well as the case where $a_r=0.05$ (recall that the total number of samples is fixed so a higher $a_r$ means fewer normal data) or learns a more complex underlying distribution that aims to jointly model both normal and anomalous samples.

The learned manifolds are visually represented in \Cref{fig: Wing_manifold_case_2} for the case of $a_r = 0.20$. This figure demonstrates the effectiveness of both LMGP-based and AE-based anomaly detection methods in encoding anomalous samples sufficiently far from the normal data in their respective manifolds. This distinction is particularly noticeable in LMGP's manifold where most of the normal data are clustered around the origin while anomalies are scattered around that cluster. 
While these manifolds (especially that of LMGP) are quite effective in separating normal and abnormal samples, \kmeans clustering incorrectly classifies some of them. For instance, in the manifold of LMGP in \Cref{fig: Wing_manifold_case_2} there is a clear gap between the orange circles (i.e., false negatives or anomalies which are incorrectly classified as normal) and the green circles (i.e., true negatives). That is, \kmeans clustering is expected to label the former as anomalies (i.e., we expect to see orange triangles instead of orange circles) but fails to do so. 

\begin{figure}[!h]
    \centering
        \includegraphics[width=1\linewidth]{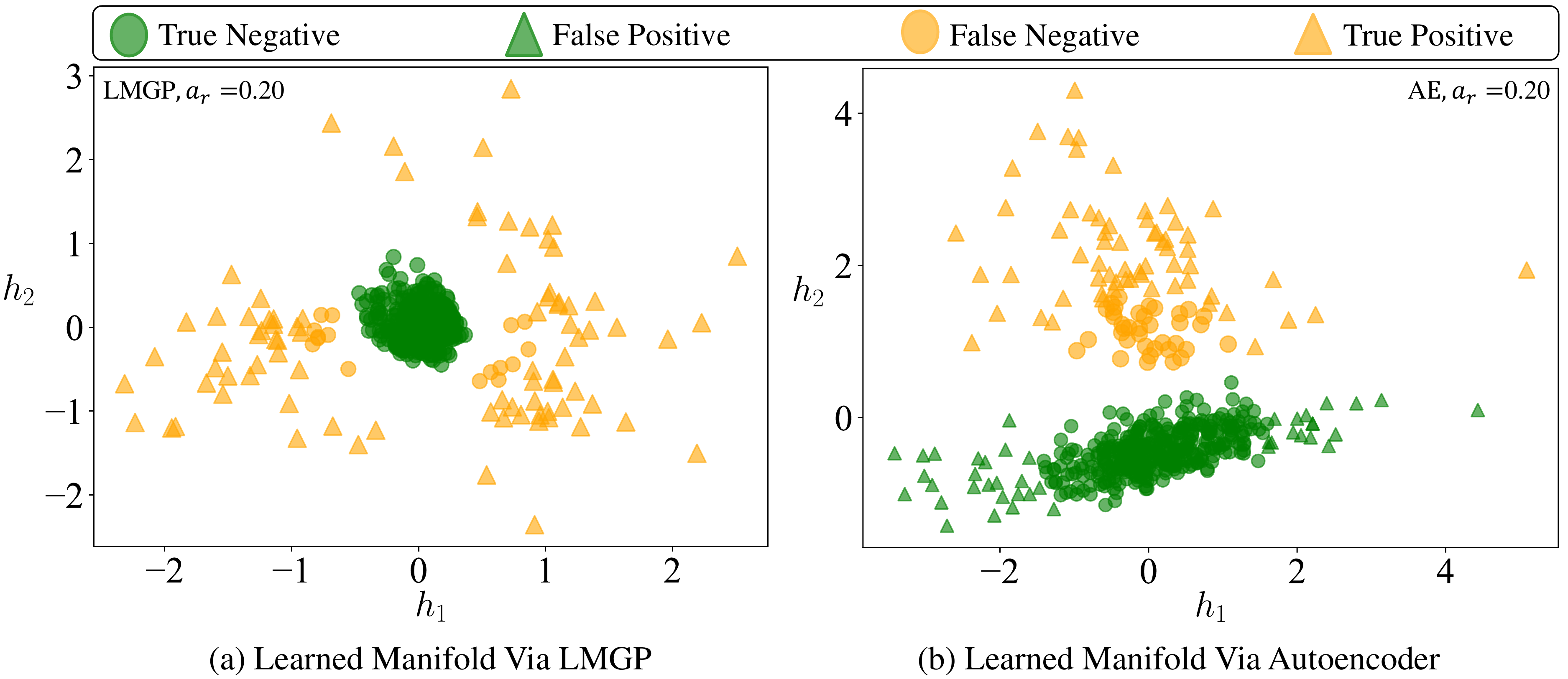}
        \vspace{-.5cm}
    \caption{\textbf{Learned manifold in the Wing model with one anomaly source:} Both approaches aim to cluster anomalous and normal samples in different clusters. In an ideal scenario, normal and anomalous samples should be denoted -via green circles and orange triangles, respectively. However, both methods fail to achieve this ideal scenario (recall that colors denote ground truth while the shapes indicate the results of our anomaly detection methods). In the case of LMGP, this error is mostly due to \kmeans clustering (see the main text for explanations) but in the case of AE it is due to both the manifold learning mechanism and \kmeans clustering.}
    \label{fig: Wing_manifold_case_2}
\end{figure}

\subsubsection{Borehole Model} \label{sec: Borehole_Example_1}
In this example, we employ the Borehole function \cite{morris1993bayesian} in \Cref{sec: appendix-equation} to generate samples and use \Cref{eq: anomaly generator} to corrupt some of these samples. \Cref{fig: Borehole_results} demonstrates the anomaly detection results for various values of $a_r$ in which LMGP consistently outperforms all other methods and is followed by our AE-based approach. For the most part, the trends in this figure are quite similar to those in \Cref{fig: Wing_results_case_2} where the $F_1$ score of isolation forest and DAGMM increase as $a_r$ increases, the performance of our LMGP-based approach slightly fluctuates and has a maximum at $a_r=0.1$, or the consistency of our AE- and LMGP-based approaches across the $20$ repetitions increases as $a_r$ increases.

\begin{figure}[!b]
    \centering
    \includegraphics[width=1\linewidth]{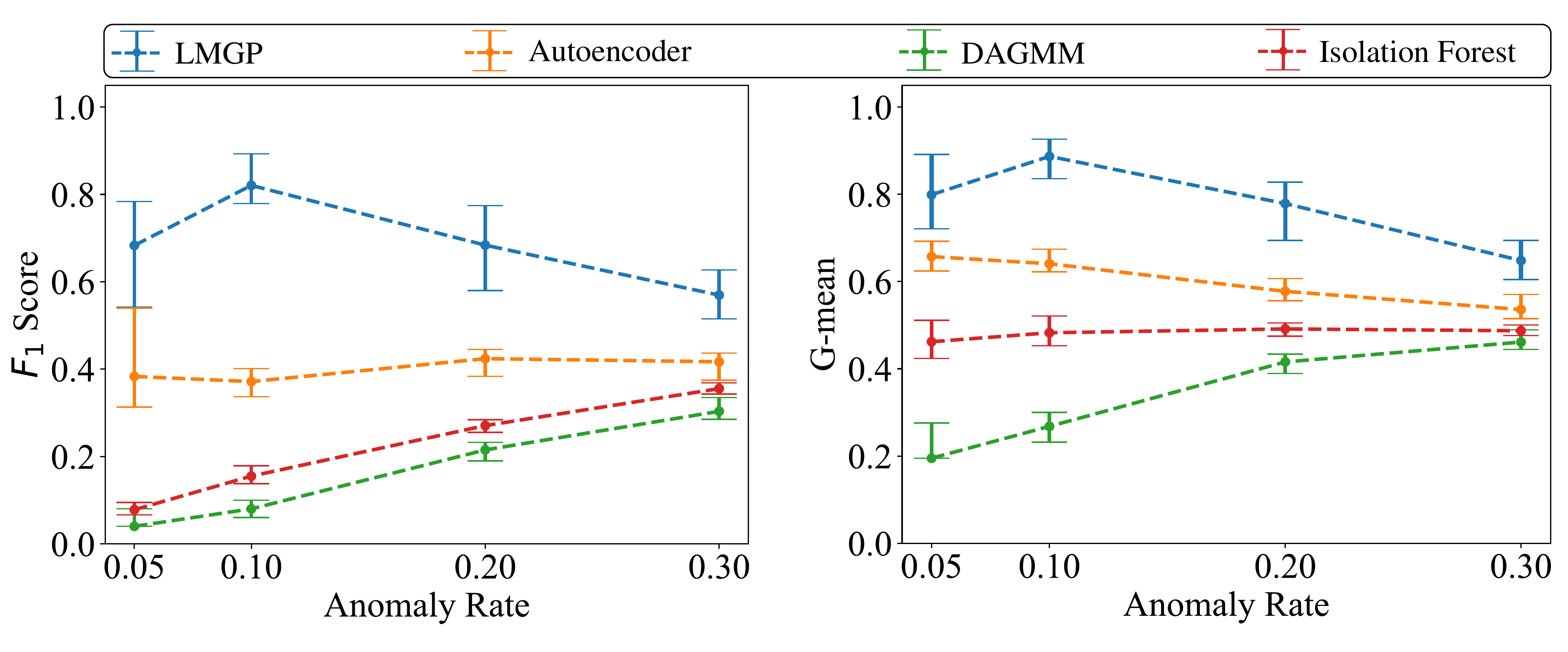}
    \vspace{-.7cm}
    \caption{\textbf{Anomaly detection results in the Borehole model:} Both LMPG and AE show superior performance. The performance of DAGMM  improves as $a_r$ increases since it leverages anomaly rate information. Also, increasing $a_r$ improves the $F_1$ value of isolation forest algorithm. This trend is because isolation forest classifies the dataset into two classes of approximately the same size regardless of the anomaly rate. As $a_r$ increases, the proportion of anomalies in the dataset grows which improves $F_1$ score as the number of false positives decreases. However, the G-mean of isolation forest negligibly changes due to the trade-off between true positive and true negative rates.}
    \label{fig: Borehole_results}
\end{figure}

However, despite the fact that the anomalies in the current and previous example are generated by the same mechanism in \Cref{eq: anomaly generator}, notable differences can be observed in the results. In particular, while our LMGP- and AE-based approaches consistently outperform isolation forest and DAGMM in this example, their performance is dropped compared to the Wing model in \Cref{sec: Wing_Example_2}. We attribute this performance drop to the fact that the input-output relation in the $8D$ Borehole model is more complex than that in the $10D$ Wing model which makes it difficult for our LMGP and AE to effectively encode samples into their manifolds such that normal and anomalous data are separated. To validate this assertion, we compare the reconstruction errors (in terms of relative root mean square error or RRMSE) of two AEs that are fitted exclusively to the normal samples in the Borehole and Wing examples. The RRMSEs in the Borehole and Wing examples are, respectively, $0.274$ and $0.1666$ which indicate that it is more difficult to learn the input-output relation in the Borehole model (in other words, more normal samples are needed in the Borehole example to match the scores achieved in the Wing example).

To gain more insights into the performance drop compared to the previous exmaple, we visualize the learnt manifolds of LMGP and AE in \Cref{fig: Borehole_learned_manifold}. Compared to \Cref{fig: Wing_manifold_case_2}, the distinction between the normal and anomalous samples is not as clear (especially in the case of AE). That is, in this example, the errors can be attributed to both \kmeans clustering and the manifold learning algorithm. To show the latter source of error in LMGP, we see in \Cref{fig: Borehole_learned_manifold} that some anomalies are incorrectly encoded close to the origin (see the orange circles close to or in the cluster of green circles) or some normal samples are encoded far from the origin (see the green triangle that is close to the bottom right corner of the plot). In both of these cases, \kmeans clustering incorrectly classifies the samples. These two error sources increase in our AE-based approach primarily because it is unable to learn an effective decision boundary that separates the normal from anomalous data.

\begin{figure}[!h]
    \centering
    \includegraphics[width=1\linewidth]{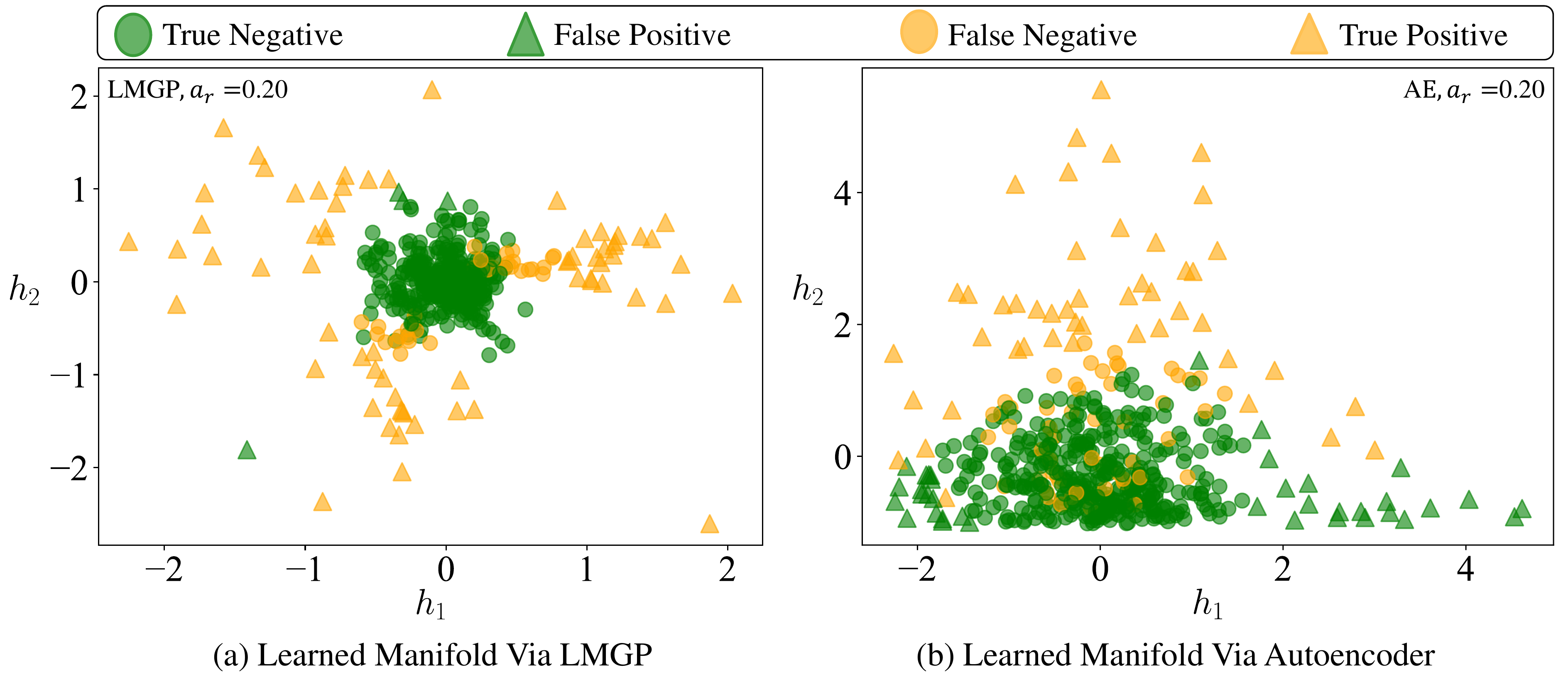}
    \vspace{-.5cm}
    \caption{\textbf{Learned manifold in the Borehole model:} Both LMGP and AE aim to differentiate between the majority of normal and anomalous samples. Upon clustering, the LMGP-based approach exhibits superior performance with fewer false positives and false negatives. Neither of the manifolds in this example are as effective as those in \Cref{fig: Wing_manifold_case_2} in differentiating the normal and anomalous data.}
    \label{fig: Borehole_learned_manifold}
\end{figure}

\subsection{Real-world Dataset} \label{sec: Real-world Datasets}
In this section, we study the performance of our anomaly detection techniques on two applications where the input space has categorical features and is higher dimensional compared to the examples in \Cref{sec: Analytic Examples}. Additionally, we do not explicitly know the input-output relation in these examples.

\subsubsection{Hybrid Organic–inorganic Perovskite (HOIPs)} \label{sec: HOIP Datasets}

HOIPs are a class of materials with unique optoelectronic properties. These materials consist of a combination of organic and inorganic components and have shown promise for applications in solar cells and other electronic devices due to their low-cost fabrication and high efficiency. Ongoing research aims to improve their stability, scalability, and environmental impact \cite{egger2016hybrid}. The dataset used here is generated via three distinct sources that simulate the band gap property of HOIPs as a function of composition based on the density functional theory (DFT). These compositions are characterized via three categorical variables that have $10$, $3$, and $16$ levels (i.e., there are $480$ unique compositions in total). The major differences between the datasets generated by the three sources are their fidelity (or accuracy) and size. Specifically, dataset generated by Source $1$ is the most accurate among the three and contains $480$ samples while the datasets generated by Source $2$ and Source $3$ contain $179$ and $240$ samples, respectively, and have lower levels of accuracy compared to Source $1$. 
Following the procedure in \Cref{sec: Wing_Example_1}, we consider the samples generated by Source $1$ as normal data and those from sources $2$ and $3$ as anomalies. The composite dataset employed in this study has a total of 500 data points and is constructed by randomly selecting normal samples from Source $1$ while equally sampling anomalous instances from sources $2$ and $3$. The ratio between normal and anomalous data is controlled via $a_r$. 
The numerical outcomes obtained from the different methods are summarized in \Cref{fig: HOIP_results}. The trends observed in isolation forest, DAGMM, and LMGP are quite close to those observed in \Cref{sec: Analytic Examples}, e.g., our LMGP-based approach consistently outperforms other methods and its performance slightly fluctuates as $a_r$ increases. Our AE-based method, however, exhibits a large variance in the $F_1$ score (and Precision presented in \Cref{sec: Precision}) when $a_r = 0.05$. We explain this behavior as follows. 
The three datasets (and hence the combined one) are quite noisy (we do not know the noise variance) as the simulations are inherently stochastic. When the anomaly rate is low there are only a few anomalous points in the combined dataset and hence detecting them is sensitive to repetition as the amount of noise and correlation can change substantially in the small data sampled from either Source $2$ or Source $3$. Consequently, our AE-based method cannot learn the underlying distribution of the normal data well and as a result flags some normal points as anomalous (false positive) when the anomaly rate is low. As opposed to the $F_1$ score, the G-mean value obtained by our AE-based approach is quite insensitive to the variations across the repetitions since both true positive and true negative rates are consistently high (note that with small $a_r$ the true negative rate remains considerably high even at high false positive rates). 
 
\begin{figure}[!h]
    \centering
    \includegraphics[width=1\linewidth]{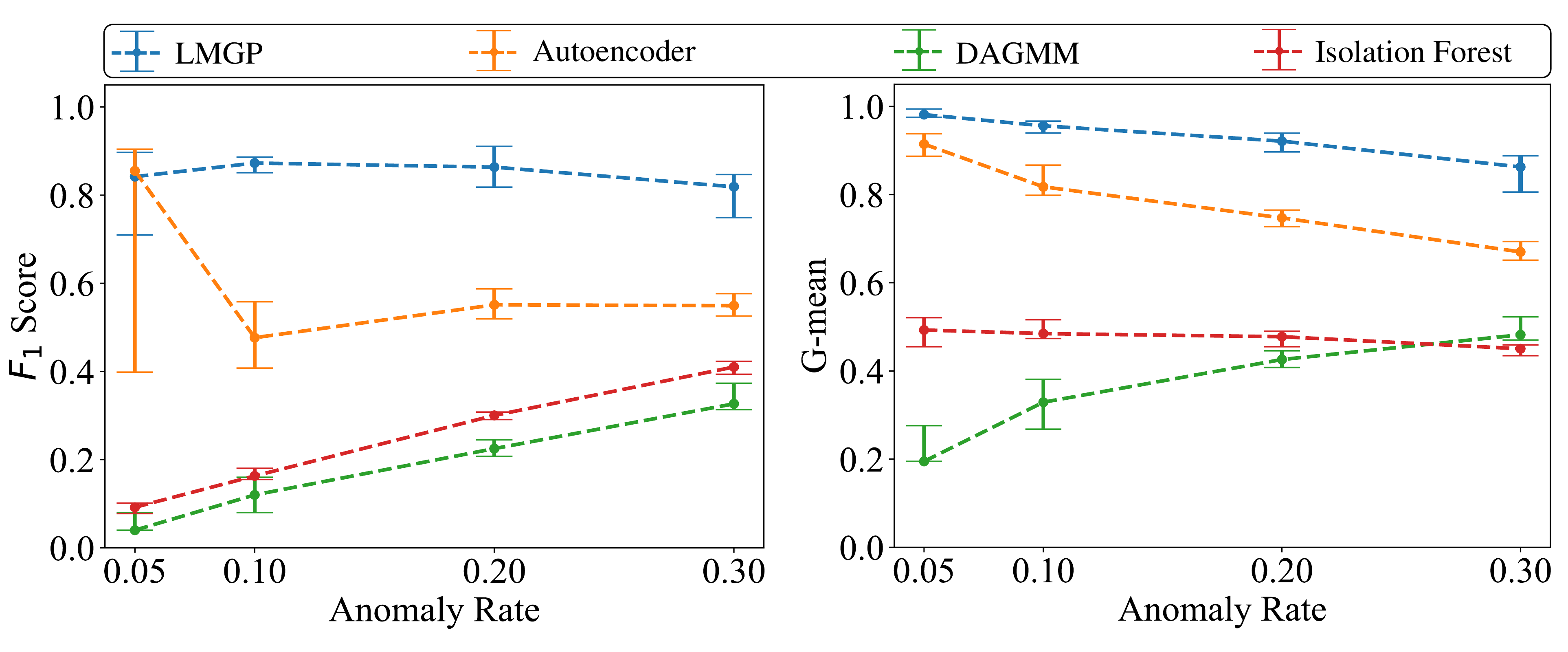}
    \vspace{-.5cm}
    \caption{\textbf{Anomaly detection results in HOIP Example:} The plots depict the results of anomaly detection techniques applied to the HOIP dataset. The large variations in $F_1$ score of AE at $a_r=0.05$ are due to noise in the dataset. AE struggles to capture the underlying distribution when there is a low anomaly rate and a high level of noise. LMGP, on the other hand, is more robust due to its modified correlation function, allowing it to handle noise and perform well.}
    \label{fig: HOIP_results}
\end{figure}

In contrast to AE, LMGP is much less sensitive to noise due to its probabilistic nature and using the so-called nugget parameter that is embedded in LMGP's kernel and directly models the noise process. Due to these features and LMGP's unique mechanism for handling categorical variables, it learnt manifold better separates normal and abnormal data, see \Cref{fig: HOIP_manifold}. In the LMGP's manifold most normal data are encoded close to the origin while in the AE's manifold these samples are encoded at the bottom part of the manifold, i.e., mostly $h_2$ is responsible for separating the normal and anomalous data. 

In \Cref{fig: HOIP_manifold} we observe that both manifolds exhibit a slight distinction between anomalous points from Source $2$ and Source $3$ where samples of Source $3$ are encoded closer to samples of Source $1$. This trend indicates that the generative distribution behind Source $1$ is closer to Source $3$ that Source $2$. Similar to previous examples, we also observe that part of the overall error is due to \kmeans clustering while the other part is due to the manifold learning. For instance, in LMGP's manifold some of the normal data are incorrectly encoded far from the origin (e.g., the green triangle towards the right edge of the plot).

\begin{figure}[!h]
    \centering
        \includegraphics[width=1\linewidth]{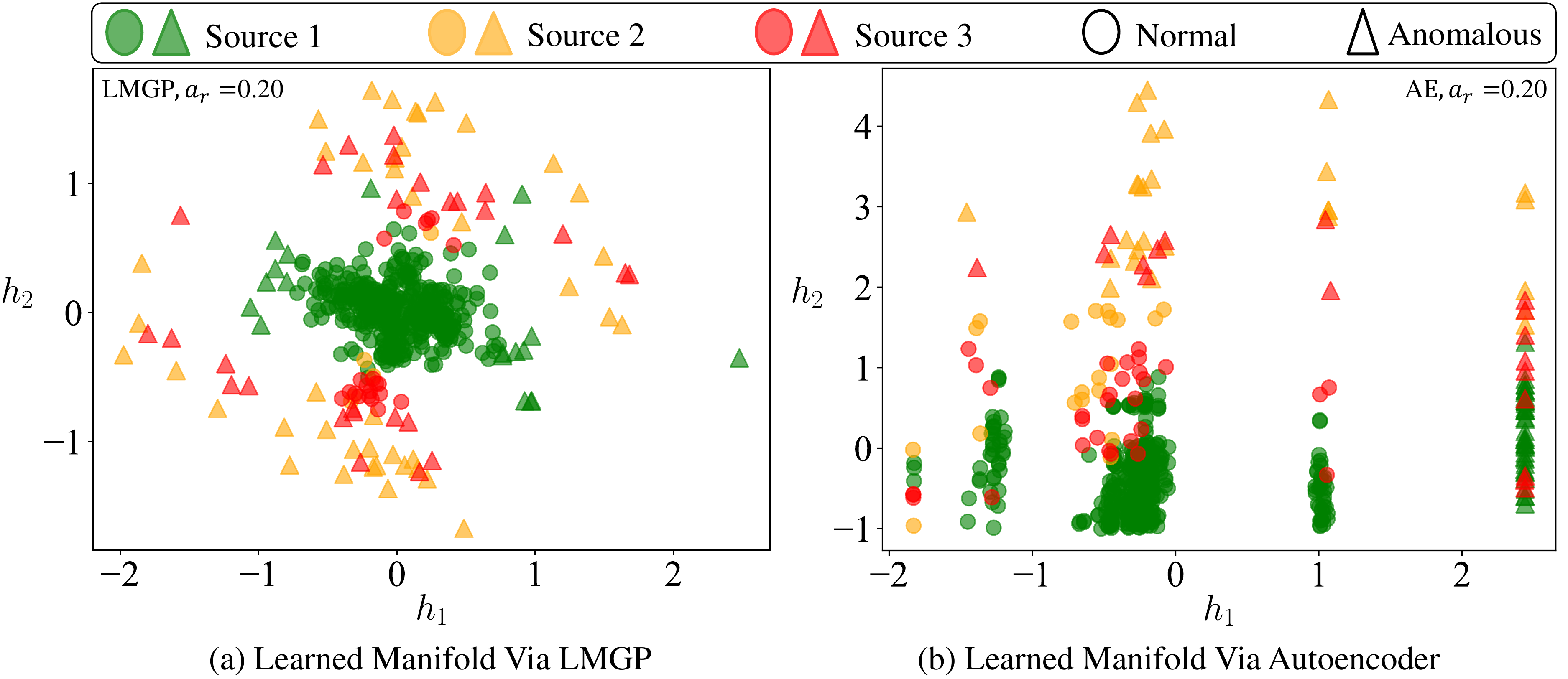}
            \vspace{-.5cm}
    \caption{\textbf{Learned manifolds in the HOIP example:} LMGP encodes the data from each source in distinct locations in the manifold. Specifically, the samples from Source $1$ are clustered close to the origin while the data points from Source $2$ are located significantly far away from the rest of the points on the manifold. The majority of the data points from Source $3$ are positioned closer to those of Source $1$ indicating that Source $1$ exhibits more similarities with Source $3$. This behavior is also observed to some extent in the manifold of AE.}
    \label{fig: HOIP_manifold}
\end{figure}

\subsubsection{High-pressure Die-casting (HPDC)}
HPDC is a widely used process for producing aluminum alloy near-net-shaped components. The process involves a machine that holds a steel die where the casting is formed and an injection system that delivers metal at high speed and holds the solidifying metal under pressure \cite{lumley2010fundamentals}. The HPDC dataset studied here has $1495$ samples where each one represents a production scenario characterized with $79$ variables (see \cite{kopper2020model} for more details). Here, we use all $1495$ samples and apply \Cref{eq: anomaly generator} to render some of the samples anomalous. 

The numerical results obtained from each method are summarized in Figure \ref{fig: HPDC_results}. The overall trends in this example are to some extent similar to those in the previous sections: our LMGP-based approach consistently outperforms other methods, the variability of our approaches at $a_r=0.05$ are relatively large due to noise and small-data issues, and the performance of DAGMM improves as $a_r$ increases. Unlike previous examples, however, the performance of isolation forest does not change as $a_r$ increases. This behavior is due to the fact that isolation forest is a neighbor-based method which suffers from curse of dimensionality, i.e., its isolation technique fails to distinguish between normal and anomalous data. This failure is due to the fact that in high dimensions and with small data the number of divisions (that this method uses to isolate anomalies) is also small for normal data. Hence, it only identifies a few samples as anomalies and, in turn, its true positive rate is close to zero regardless of the value of $a_r$.

\begin{figure}[!t]
    \centering
        \includegraphics[width=1\linewidth]{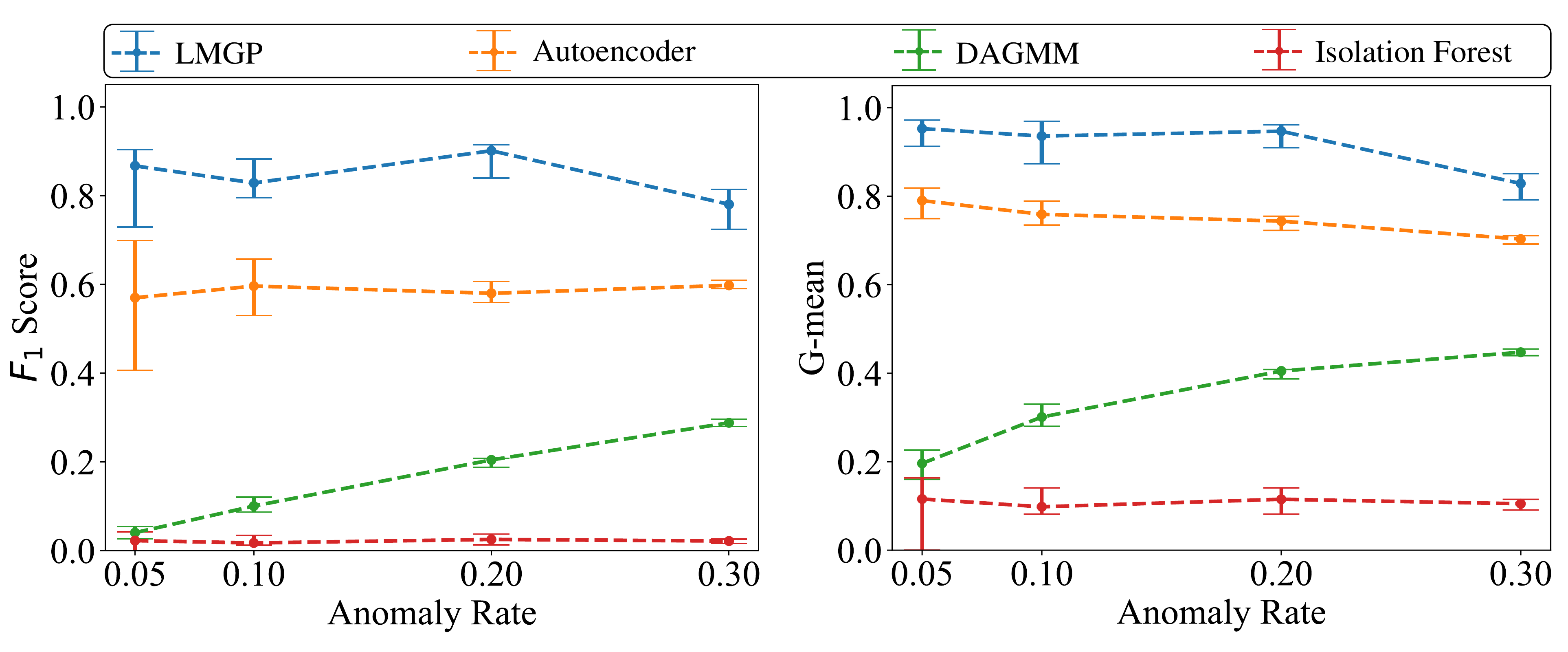}
     \vspace{-.5cm}
    \caption{\textbf{Anomaly detection results in the HPDC example:} Although the data is high dimensional, our LMGP-based approach produces quite accurate results while DAGMM and especially isolation forest perform poorly in distinguishing between normal and anomalous samples.}
    \label{fig: HPDC_results}
\end{figure}
 
\begin{figure}[!t]
    \centering
    \includegraphics[width=1\linewidth]{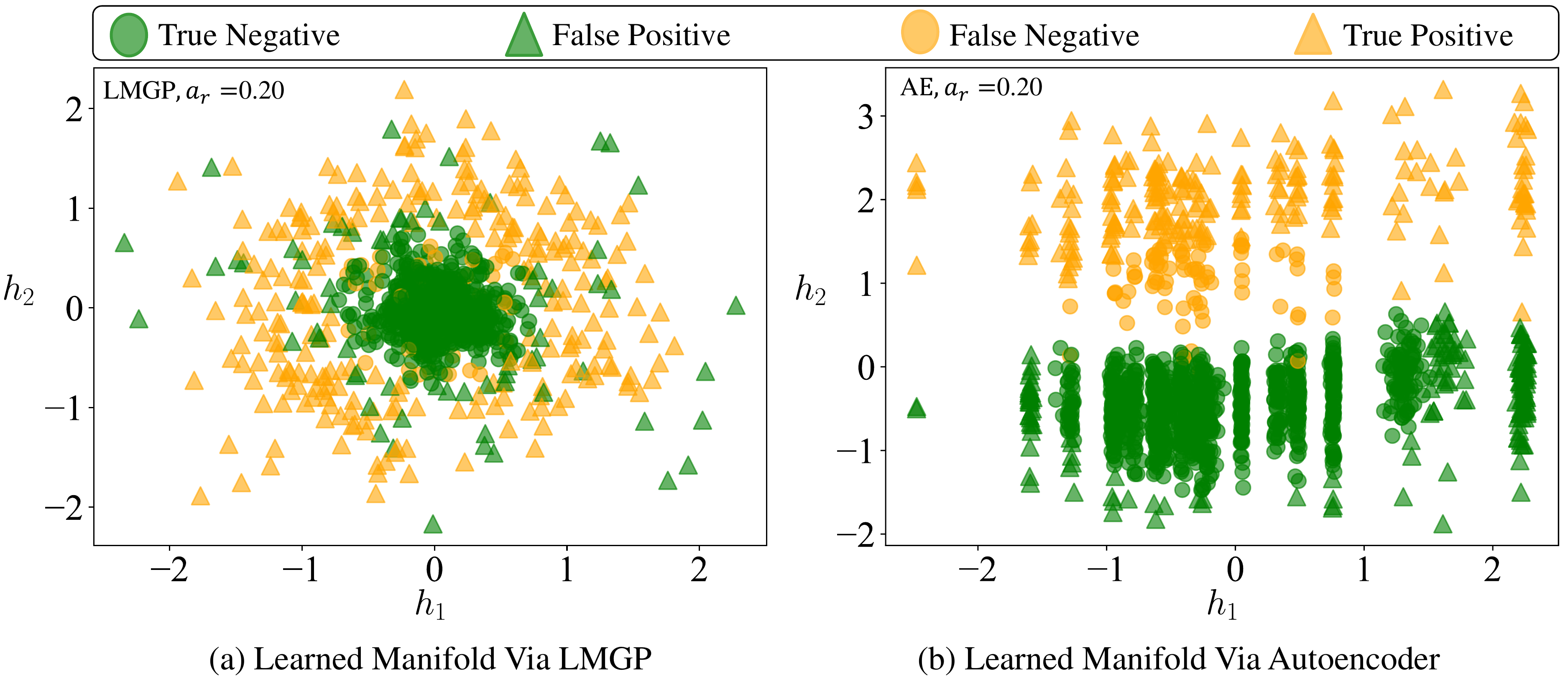}
    \vspace{-.2cm}
    \caption{\textbf{Learned manifold in HPDC example:} LMGP scatters anomalous samples across various positions without forming distinct clusters which suggests that the anomalies in this example have a random nature. In contrast, the AE-based technique solely relied on the utilization of $h_2$ to cluster anomalous and normal points and it does not provide any further insights into the characteristics or nature of the anomalous samples.}
    \label{fig: HPDC_manifold}
\end{figure}

\Cref{fig: HPDC_manifold} illustrates the learned manifolds where in the case of LMGP normal samples are mostly placed close to the origin while anomalous ones are spread out around them without forming distinct clusters. This distribution is a result of the random nature of the anomalous samples and is expected. In contrast, the AE-based method lacks the ability to uncover the inherent randomness in anomalies and primarily relies on the reconstruction error ($h_2$) to detect anomalous samples.

    \section {Conclusion} \label{sec: conclusion}
We introduce an unsupervised anomaly detection technique based on nonlinear manifold learning. Our approach has two primary stages. First, using either LMGP or AE, we embed all data points in a low-dimensional and visualizable latnet space that preserves the underlying structure of the data while separating normal samples from the anomalous ones. Then, we use the \kmean clustering algorithm to group the encoded points into clusters based on their positions on the manifold. Unlike most existing methods, our approach operates entirely in an unsupervised manner and does not require any information about the anomaly rate. We compare our method against two of the state-of-the-art techniques on three analytic and two real-world datasets with different levels of  complexity and dimensionality. The results demonstrate that our LMGP-based approach consistently outperforms other techniques and is followed by our AE-based method. 

A particularly useful feature of our LMGP is its pre-defined framework that does not require any parameter tuning which makes it easier to implement. Moreover, our results indicate that the learned manifold by LMGP not only distinguishes between anomalous and normal samples, but also has the potential to provide insights into the characteristics of anomalies (e.g., the number of different mechanisms that generate anomalous data, see \Cref{fig: HOIP_manifold} for an example). Furthermore, our LMGP-based method demonstrates a remarkable robustness against noise as it has a probabilistic nature and directly models the noise process within its kernel.

In the present study we set $k=2$ in the \kmeans clustering regardless of the obtained manifold. Developing an automated approach to dynamically adjust the value of $k$ based on the characteristics of the learned manifold has the potential to improve the results. Furthermore, in our current approach, two steps are performed sequentially. However, we anticipate that the effectiveness of our LMGP- and AE-based approaches can be further enhanced by integrating and performing these steps jointly. We aim to investigate these directions in our future works. 

\section*{Acknowledgments}
We appreciate the support from the Office of the Naval Research (award number N000142312485), Early Career Faculty grant from NASA’s Space Technology Research Grants Program (award number 80NSSC21K1809), and the UC National Laboratory Fees Research Program of the University of California (Grant Number L22CR4520).
\appendix
\addcontentsline{toc}{section}{Appendices}
\section*{Appendices}

\setcounter{equation}{0}
\renewcommand{\theequation}{\thesection-\arabic{equation}}

\section{Functional Form of the Analytic Models} \label{sec: appendix-equation}
Source 1, 2, and 3 for wing function studied in section \ref{sec: Analytic Examples} are respectively given by:
\begin{equation} 
    \begin{split}
        y(\boldsymbol{x})=0.36 s_w^{0.758} w_{f w}^{0.0035}(\frac{A}{\cos ^2(\Lambda)})^{0.6} q^{0.006} \times \lambda^{0.04}(\frac{100 t_c}{\cos (\Lambda)})^{-0.3}(N_z W_{d g})^{0.49}+s_w w_p+ \epsilon&
    \end{split}
    \label{eq: Wing_HF-function}
\end{equation}
\begin{equation} 
    \begin{split}
        y(\boldsymbol{x})=0.36 s_w^{0.8} w_{f w}^{0.0035}(\frac{A}{\cos ^2(\Lambda)})^{0.6} q^{0.006} \times \lambda^{0.04}(\frac{100 t_c}{\cos (\Lambda)})^{-0.3}(N_z W_{d g})^{0.49}+w_p+ \epsilon&
    \end{split}
    \label{eq: Wing_LF1-function}
\end{equation}
\begin{equation} 
    \begin{split}
        y(\boldsymbol{x})=0.36 s_w^{0.9} w_{f w}^{0.0035}(\frac{A}{\cos ^2(\Lambda)})^{0.6} q^{0.006} \times \lambda^{0.04}(\frac{100 t_c}{\cos (\Lambda)})^{-0.3}(N_z W_{d g})^{0.49}+ \epsilon&
    \end{split}
    \label{eq: Wing_LF2-function}
\end{equation}
where $y(\boldsymbol{x})$ represents the response and the input vector is $\boldsymbol{x}=[s_w, w_{f w}, A, \Lambda, q, \lambda, t_c, N_z, W_{d g},w_p]^T$. Also, $\epsilon \sim \mathcal N(0,5^2)$ represents the noise with zero mean and standard deviation of $5$. The noise variance is defined based on the range of each function. For more information on these models and their accuracy with respect to Source $1$ we refer the reader to \cite{eweis2022data}. 

The analytic Borehole function studied in section \ref{sec: Analytic Examples} is given by:
\begin{equation} 
    \begin{split}
        y(\boldsymbol{x})=\frac{2 \pi T_{u}(H_u-H_l)}{\ln (\frac{r}{r_w})(1+\frac{2 L T_u}{\ln (\frac{r}{r w}) r_w^2 k_w}+\frac{T_{u}}{T_l})}+\epsilon&
    \end{split}
    \label{eq: Borehole-function}
\end{equation}
where the input vector is $\boldsymbol{x}=[T_{u}, H_u, H_l, r, {r_w},L, k_w, T_l]^T$. Also, $\epsilon \sim \mathcal N(0,3.40^2)$ represents the noise with zero mean and standard deviation of $3.40$.

\section{Metrics} \label{sec: Metrics definition}
We choose three metrics in our evaluations, namely
F1-Score, Precision, and G-mean. The formulas of F1-Score and precision are given by: 
\begin{equation} 
    \begin{split}
        \mathrm{F1-score=\frac{TP}{TP+0.5(FP+FN)}}
    \end{split}
    \label{eq: F1-score&Precision}
\end{equation}
\begin{equation} 
    \begin{split}
        \mathrm{Precision=\frac{TP}{TP+FP}}
    \end{split}
    \label{eq: F1-score&Precision}
\end{equation}
where TN and TP denote the number of True Negatives and True Positives, respectively. Also, FN and FP indicate the number of False Negatives and False Positives, respectively. In addition,  G-mean is defined as:
\begin{equation} 
    \begin{split}
         \mathrm{{G-mean=\sqrt{\frac{TP}{TP + FN}\times \frac{TN}{TN + FP}}}}
    \end{split}
    \label{eq: G-mean}
\end{equation}

\section{Precision} \label{sec: Precision}
Here we present precision results across all examples. By applying various anomaly detection techniques to all examples, the precision results are visually represented in Figure \ref{fig: Precision_results}. Our LMGP- and AE-based methods outperform all other techniques tested.

\begin{figure}[H]
    \centering
        \includegraphics[width=1\linewidth]{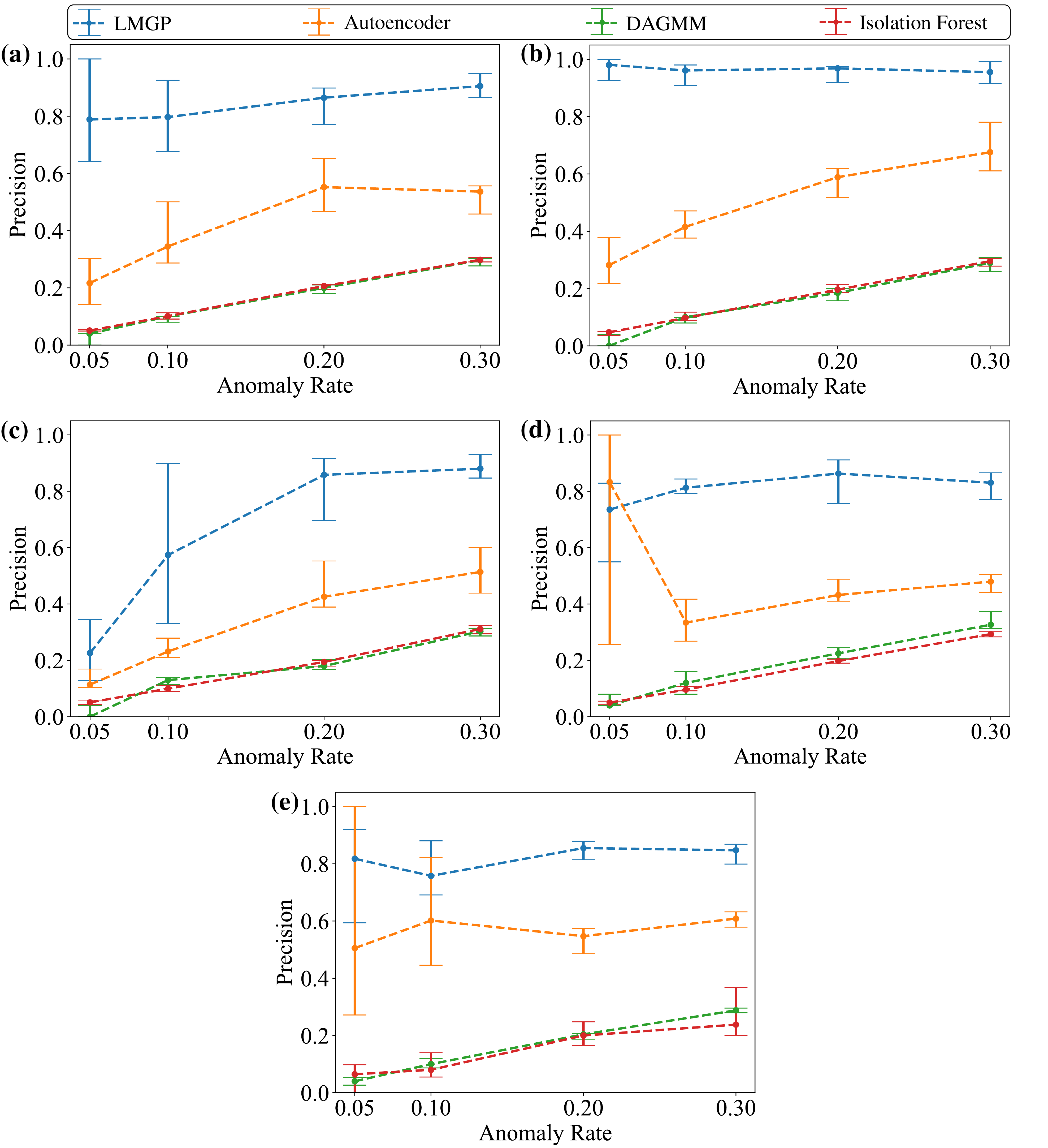}
        \vspace{-.5cm}
    \caption{\textbf{Precision results in all examples:} The plots illustrate the precision results obtained by applying various anomaly detection techniques to all examples. Both LMG-based and AE-based methods outperform all other techniques while the LMG-based method proposed in this study achieves the highest rank among all techniques.}
    \label{fig: Precision_results}
\end{figure}

    \pagebreak    
    \printbibliography
\end{document}